\documentclass{article}
\PassOptionsToPackage{numbers,sort&compress}{natbib}
\usepackage[letterpaper,textwidth=5.5in,textheight=9in,centering]{geometry}
\usepackage{times,natbib}
\setcitestyle{numbers,square,citesep={,},aysep={,},yysep={,}}
\ifdefined\XeTeXversion
\usepackage{fontspec}
\fi
\usepackage{amsmath,amssymb,booktabs,graphicx}
\usepackage[hidelinks]{hyperref}
\usepackage{url}
\usepackage{flafter}\usepackage{float} % Tables cannot float before their source position.
\title{Generative Retrieval for\\Reverse Target Fishing\\with a Proteome-Scale Benchmark}
\author{\begin{tabular}{c}
Zhichao Deng\textsuperscript{1,2,3,*,\(\ddagger\)},
Haichuan Tan\textsuperscript{1,2,4,*},
Xin Hong\textsuperscript{2},\\
Huaizhi Wang\textsuperscript{1,2,3},
Tao Huang\textsuperscript{1,2,3},
Bowen Gao\textsuperscript{1,2,4},\\
Yanyan Lan\textsuperscript{1,2,\(\dagger\)}\\[6pt]
\small \textsuperscript{1}AIPher Technology\\
\small \textsuperscript{2}Institute for AI Industry Research (AIR), Tsinghua University\\
\small \textsuperscript{3}School of Pharmaceutical Sciences, Peking University\\
\small \textsuperscript{4}Department of Computer Science and Technology, Tsinghua University\\[4pt]
\small Zhichao Deng: \texttt{2310307113@stu.pku.edu.cn}\\[2pt]
\small \textsuperscript{*}These authors contributed equally to this work.\\
\small \textsuperscript{\(\dagger\)}Corresponding author: \texttt{lanyanyan@air.tsinghua.edu.cn}\\
\small \textsuperscript{\(\ddagger\)}Work done during an internship at AIPher Technology.
\end{tabular}}

\date{}
\newcommand{\sg}{\operatorname{sg}}
\newcommand{\mse}{\operatorname{MSE}}
\begin{document}
\maketitle
\begin{abstract}
Reverse target fishing identifies protein targets for a query molecule and is central to drug discovery and repurposing. Discriminative approaches adapted from virtual screening face sparse, long-tailed associations and missing-not-at-random measurements: treating unrecorded pairs as negatives can suppress genuine targets. We reformulate the task as generative retrieval and introduce \textbf{GenFish}. Residual vector quantization converts frozen protein language model features into shared semantic identifiers. Given a molecular fingerprint, GenFish autoregressively models codes conditioned on the query and preceding code, ranking proteins by identifier likelihood without explicitly labeling unrecorded pairs as negatives. We introduce \textbf{ProTargetBench}, the first proteome-scale benchmark for reverse target fishing to jointly support measured-panel discrimination and full-catalogue recovery of experimentally recorded positives. Covering 20,431 UniProt-reviewed human proteins, it integrates curated bioactivity evidence and distinguishes experimental inactivity from unmeasured associations. On 198 held-out molecules, GenFish achieves higher mean performance across all three full-catalogue retrieval metrics than four discriminative neural baselines and three non-neural baselines, including ligand-based methods. A controlled positive-hiding experiment shows that targeted removal of false-negative competition improves contrastive retrieval beyond matched random masking; GenFish remains stronger under the same incomplete supervision. Pfam alignment indicates that the semantic identifiers capture biologically meaningful relationships among proteins. These results establish generative retrieval as a principled paradigm for reverse target fishing under incomplete bioactivity records.

\end{abstract}

\section{Introduction}

Reverse target fishing asks which proteins in the human body a small molecule may
bind to or affect. Identifying these proteins can help us understand how a drug
works, detect unintended interactions, and discover new uses for existing
compounds \citep{keiser2009,swisstarget2019}. Unlike conventional virtual
screening, which fixes a protein and searches for molecules, reverse target
fishing fixes the molecule and searches over a large protein catalogue. The
task is therefore to rank many possible protein targets for each molecule.

\begin{figure}[!htb]
\centering
\includegraphics[page=1,trim=70 55 35 55,clip,width=\linewidth]{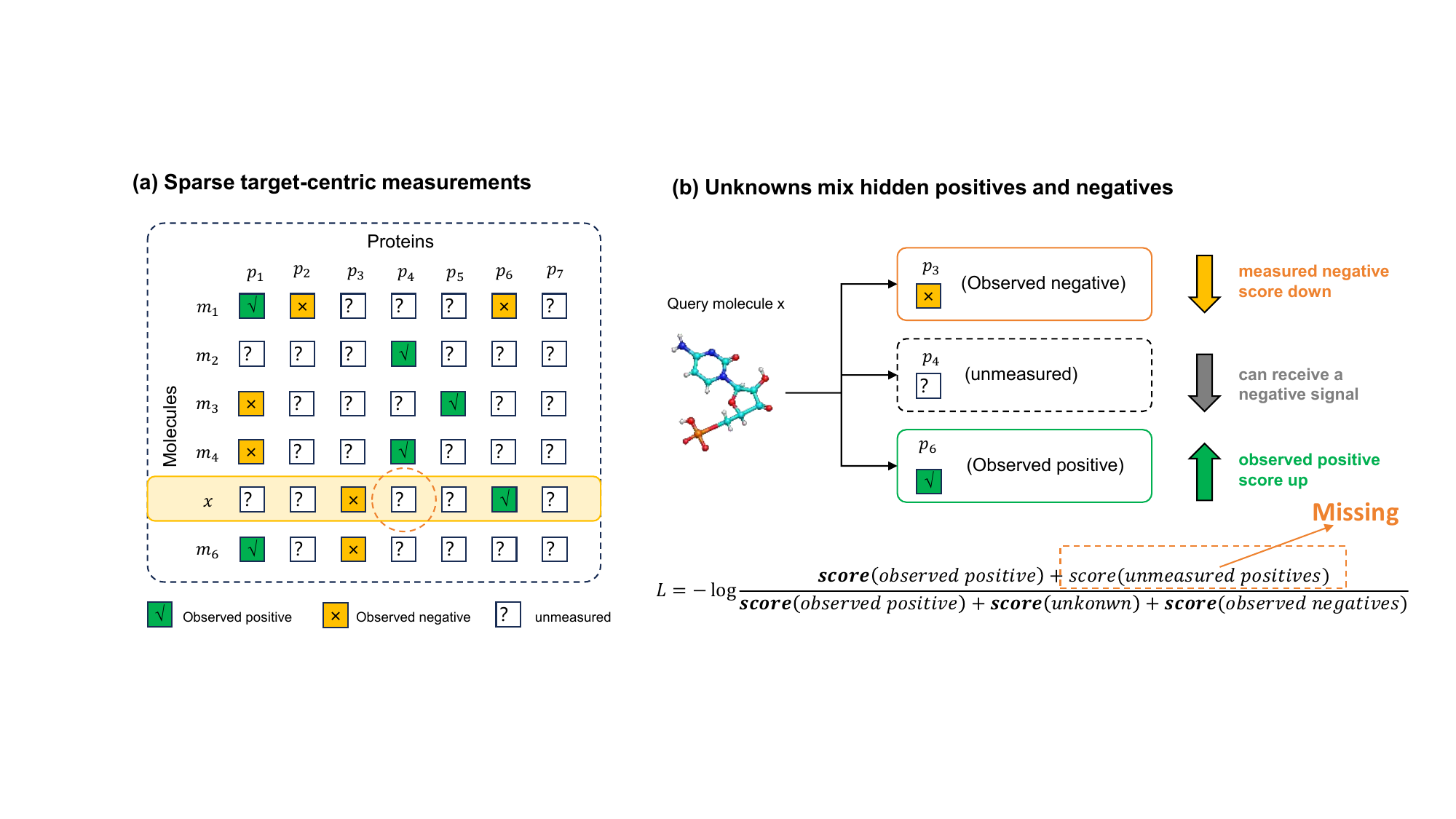}
\caption{\textbf{Why sparse reverse target-fishing supervision is difficult.} Unmeasured molecule--protein pairs may include both hidden binders and non-binders; treating candidate-level unknowns as negative competitors can suppress undiscovered targets.}
\label{fig:motivation}
\end{figure}

A central difficulty is that bioactivity databases record molecule--protein
associations selectively. An observed pair can provide positive activity
evidence, whereas an unrecorded pair may simply not have been measured. This
distinction is especially important because assays focus on selected proteins,
leaving individual molecules sparsely annotated and targets unevenly studied.
Our training data contain 381,225 positive associations for 291,337 molecules,
or only about 1.3 recorded targets per molecule on average. Treating all
unrecorded pairs as negative samples or contrastive competitors can therefore
suppress genuine but undiscovered targets, as illustrated in
Figure~\ref{fig:motivation}.

This asymmetry raises a modelling question: can sparse positive associations
share information across proteins without assigning unknown molecule--protein
pairs negative labels? A useful representation should transfer supervision
across related proteins while preserving distinctions between candidates. We
address this question with \textbf{GenFish}: protein representations are
quantized into shared, multi-level semantic identifiers, allowing observed
associations to share supervision across candidates without explicitly treating
unmeasured pairs as negatives \citep{dsi2022,tiger2023}.

Specifically, GenFish uses residual vector quantization to construct semantic
identifiers from frozen protein language-model features. Given a molecule, it
autoregressively predicts the identifier codes and ranks the fixed protein
catalogue by their likelihood, using molecule-conditioned transitions between successive semantic codes. This preserves molecule-specific preferences while sharing
statistical strength across related proteins.

Evaluating this problem requires a benchmark that covers both capabilities.
Existing evaluations either discriminate activity within selected assay panels
or score candidates in a broad structural catalogue: the former lacks
proteome-scale recovery, while the latter usually lacks curated supervision
and does not distinguish experimental inactivity from unmeasured pairs. We
therefore introduce \textbf{ProTargetBench}, which integrates curated
associations from ChEMBL, BindingDB, and PubChem
\citep{chembl2019,bindingdb2025,pubchem2025} with a fixed catalogue of 20,431
UniProt-reviewed human proteins. It evaluates measured-panel discrimination
and full-catalogue recovery under the same candidate space while preserving the
distinction between inactive and unmeasured pairs.

On 198 held-out molecules, GenFish achieves the highest mean performance among
the evaluated neural and non-neural baselines on all three full-catalogue
retrieval metrics. It reaches 35.27\% R@10, 51.96\% R@50, and 46.31\% MRR,
including a 5.46-point R@50 improvement over the strongest neural comparator.
A controlled positive-hiding experiment shows that removing false-negative
competition improves contrastive retrieval, while GenFish remains stronger
under the same incomplete supervision. Pfam analysis shows that the identifiers retain
biologically meaningful structure.

Our contributions are:
\begin{itemize}
\item \textbf{Method:} We formulate reverse target fishing as retrieval over
protein semantic identifiers and introduce \textbf{GenFish}, which learns from positive
associations without explicitly treating unrecorded pairs as negatives.

\item \textbf{Benchmark:} We introduce \textbf{ProTargetBench}, a proteome-scale
benchmark spanning 20,431 UniProt-reviewed human proteins and separating
measured-panel discrimination from full-catalogue positive recovery.

\item \textbf{Evidence:} GenFish outperforms the evaluated neural and
non-neural baselines on full-catalogue retrieval, while controlled experiments
demonstrate the effect of false-negative competition and the contributions of
the proposed identifier structure.
\end{itemize}

\section{Related Work}

\textbf{Target fishing and molecular--protein retrieval.}
Reverse target-fishing methods include ligand-based similarity, structure-based reverse docking, and neural molecule--protein matching. SEA and SwissTargetPrediction transfer information from known active compounds \citep{sea2007,swisstarget2019}, while reverse docking and structure-aware models score candidate binding sites \citep{tarfisdock2006,maitargetfisher2025}. Neural methods learn molecular and protein representations for interaction or affinity prediction, including contrastive retrieval models such as ConPLex and DrugCLIP and graph- or geometry-aware models such as LigUnity and HypSeek \citep{deepdta2018,graphdta2021,moltrans2021,conplex2023,drugclip2023,ligunity2025,hypseek2025}. These approaches generally score molecule--protein pairs discriminatively, which can treat unmeasured pairs as negatives. Existing evaluations emphasize either measured-panel discrimination or broad structural candidate sets \citep{papyrus2023,mayr2018,peon2016,beta2022,luo2024reverse}; they do not combine a fixed proteome-scale catalogue with explicit positive, inactive, and unmeasured states. ProTargetBench addresses this gap by combining a fixed proteome-scale catalogue with measured-panel discrimination and full-catalogue recovery of recorded positives.

\textbf{Generative retrieval and semantic identifiers.}
Generative retrieval models candidate identifiers conditioned on a query. DSI and TIGER established this paradigm for document retrieval and recommendation using semantic identifiers and residual quantization \citep{dsi2022,tiger2023}; LETTER studied related identifier structures \citep{letter2024}. GenFish adapts these ideas to protein retrieval by learning molecule-conditioned identifier likelihoods from observed positives.
\section{Problem Formulation and GenFish}
\label{sec:method}
\label{sec:task}

\subsection{Problem Formulation}

Let $\mathcal{X}$ denote the space of query molecules and let
$\mathcal{P}$ be a fixed catalogue of candidate proteins. Given a query
molecule $x\in\mathcal{X}$, reverse target fishing aims to rank proteins
$p\in\mathcal{P}$ according to whether they may be affected by $x$. A molecule
may have multiple relevant proteins, so the task is a multi-positive retrieval
problem rather than single-label classification.

The available training data consist of observed positive associations
\[
\mathcal{D}^{+}=\{(x_i,p_i)\}_{i=1}^{N}.
\]
An unrecorded pair $(x,p)\notin\mathcal{D}^{+}$ is unknown: it may correspond
to a non-binder or to an interaction that has not been measured. We therefore
do not define a complementary negative set over
$\mathcal{X}\times\mathcal{P}$. Instead, the learning objective uses observed
positive associations while leaving unrecorded pairs unlabeled.

The goal is to learn a query-dependent ranking function
$s_\theta:\mathcal{X}\times\mathcal{P}\rightarrow\mathbb{R}$ and return
\[
\operatorname{TopK}_\theta(x;\mathcal{P})
=
\underset{\substack{S\subseteq\mathcal{P}\,\,|S|=K}}{\arg\max}
\sum_{p\in S}s_\theta(x,p).
\]
The catalogue defines the search space, while the recorded positives define the
known targets available for training and evaluation. GenFish is designed to
mitigate sparse supervision by sharing statistical strength across protein
identifiers without assigning unmeasured molecule--protein pairs negative
labels.

\subsection{GenFish Overview}

\begin{figure}[!htb]
\centering\includegraphics[width=\linewidth]{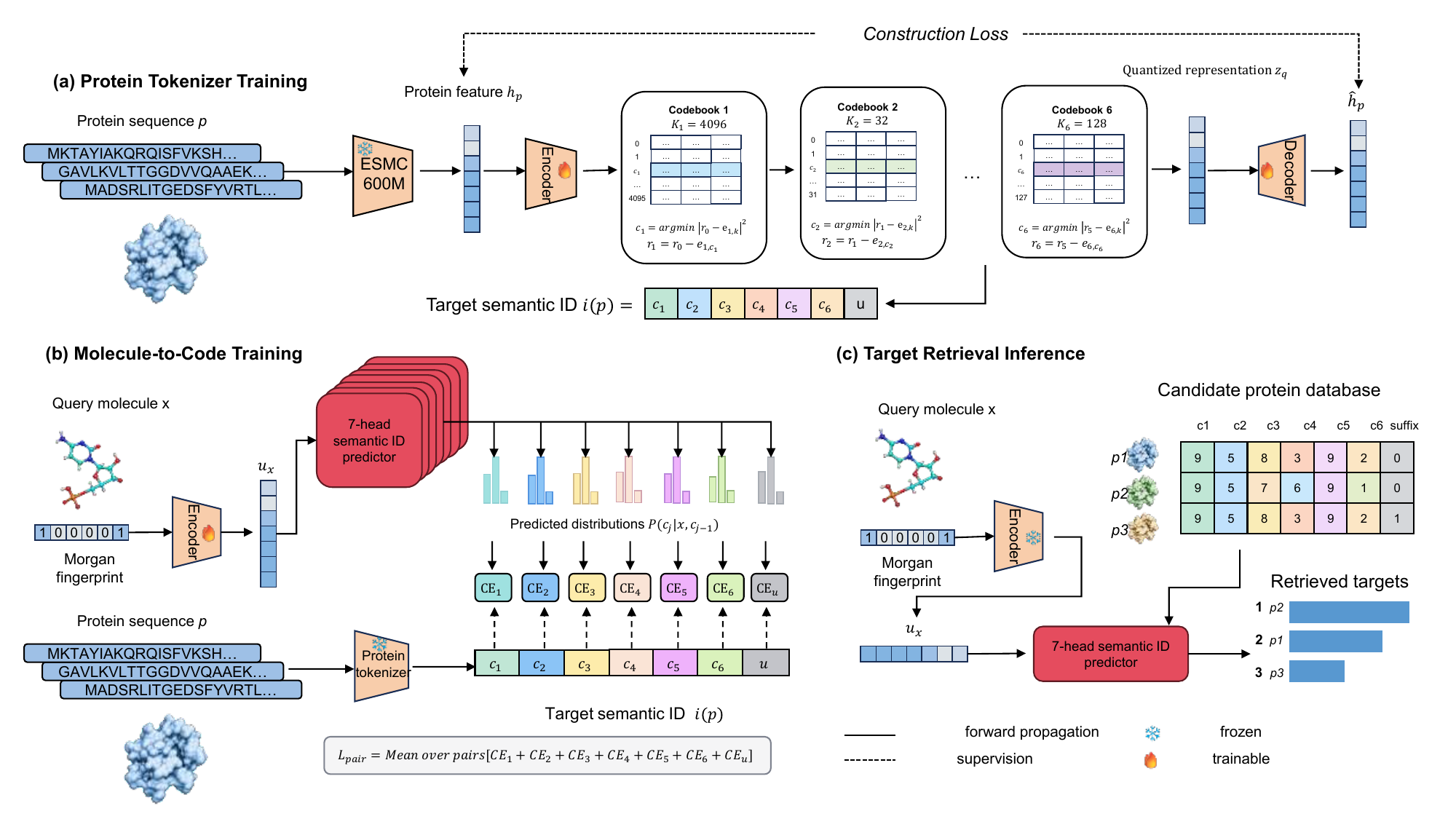}
\caption{\textbf{Overview of GenFish.}
(a) Protein tokenization: seven-component protein identifiers are constructed
from quantized frozen ESMC-600M features.
(b) Molecular prediction: observed positive associations train a molecular
predictor to assign likelihoods to target identifiers from Morgan fingerprints.
(c) Catalogue retrieval: candidate proteins are ranked by identifier
likelihood, with each subsequent semantic code conditioned on the query
molecule and the preceding code.
Code values are illustrative and do not imply numerical similarity.}
\label{fig:genfish-overview}
\end{figure}

GenFish represents each protein with a discrete semantic identifier derived from
its sequence representation. The identifier is constructed from shared
codebooks, allowing related proteins to share identifier components. A
molecule-conditioned predictor then assigns likelihood to candidate
identifiers. At inference, every catalogue protein is scored using its stored
identifier, and candidates are ranked by identifier likelihood
(Figure~\ref{fig:genfish-overview}).

\subsection{Protein Semantic Identifiers}

We extract sequence features using frozen ESMC-600M and standardize them to
obtain a protein representation $\mathbf h_p$. An encoder maps this
representation into a low-dimensional latent space,
\[
\mathbf z_p=E(\mathbf h_p).
\]
Residual vector quantization converts the latent representation into a sequence
of discrete codes. At each level, the nearest codebook vector is selected and
subtracted from the remaining residual:
\begin{equation}
\mathbf r_0=\mathbf z_p,\qquad
c_j(p)=\arg\min_k\|\mathbf r_{j-1}-\mathbf e_{j,k}\|_2^2,\qquad
\mathbf r_j=\mathbf r_{j-1}-\mathbf e_{j,c_j(p)}.
\label{eq:id-quantization}
\end{equation}
Here $\mathbf e_{j,k}$ denotes the $k$-th vector in the level-$j$ codebook. A
decoder reconstructs the protein features from the sum of the selected vectors.
The encoder and codebooks are trained using reconstruction and quantization
objectives, without molecule--protein interaction labels.

GenFish represents each protein with a seven-code identifier
\[
\mathbf c(p)=\big(c_1(p),\ldots,c_6(p),c_7(p)\big).
\]
Once tokenization is complete, all protein-side parameters and identifiers are
frozen. Proteins that share code components also share the corresponding output
parameters in the molecular predictor, allowing supervision from observed
associations to propagate across related candidates.

\subsection{Molecule-Conditioned Identifier Prediction}

For a query molecule $x$, a multilayer perceptron maps its Morgan fingerprint to
a query representation $\mathbf u_x$. The base logits at code level $j$ are
\[
\mathbf a_j(x)=W_j\mathbf u_x+\mathbf b_j.
\]
For the intermediate levels $j=2,\ldots,6$, GenFish adds a low-rank interaction
conditioned on the preceding code. Given preceding code $a$, the logit for
current code $k$ is
\begin{equation}
\ell_{j,k}(x,a)
=
a_{j,k}(x)
+
[\mathbf g_j(x)\odot\bar{\mathbf U}_j[a]]^\top
\mathbf V_j[k],
\qquad
\mathbf g_j(x)=G_j\mathbf u_x+\mathbf d_j,
\label{eq:conditional-logit}
\end{equation}
where $\bar{\mathbf U}_j[a]$ is the normalized embedding of the preceding code,
$\mathbf V_j[k]$ is the current-code embedding, and $\odot$ denotes
elementwise multiplication. The molecular gate allows the same preceding code
to induce different predictions for different query molecules.

For the first and final code levels, the preceding-code interaction is omitted.
Thus all seven code-level distributions can be written
in the unified form
\begin{equation}
P_{\theta,j}(k\mid x,a)
=
\frac{\exp\ell_{j,k}(x,a)}
{\sum_{v=0}^{K_j-1}\exp\ell_{j,v}(x,a)},
\qquad j=1,\ldots,7,
\label{eq:conditional-prob}
\end{equation}
where $a=c_{j-1}(p)$ for $j=2,\ldots,6$, while $a=\varnothing$ for $j=1$
and $j=7$. For $j=1$ and $j=7$, the resulting distribution depends only on
$x$.

Let $c_0(p)=\varnothing$. The complete identifier likelihood is then
\begin{equation}
P_\theta(\mathbf c(p)\mid x)
=
\prod_{j=1}^{7}
P_{\theta,j}\!\left(c_j(p)\mid x,c_{j-1}(p)\right).
\label{eq:identifier-likelihood}
\end{equation}
This factorization shares parameters across proteins while allowing the query
molecule to modulate the compatibility of successive identifier components.

\subsection{Training and Catalogue Retrieval}

Given the observed positive associations $\mathcal D^+$, GenFish minimizes the
negative log-likelihood of the stored target identifiers:
\begin{equation}
\mathcal L(\theta)
=
-\frac{1}{|\mathcal D^+|}
\sum_{(x,p)\in\mathcal D^+}
\sum_{j=1}^{7}
\log P_{\theta,j}\!\left(c_j(p)\mid x,c_{j-1}(p)\right).
\label{eq:pair}
\end{equation}
Each positive association contributes one training example. Multiple recorded
targets for the same molecule therefore provide multi-positive supervision,
while unrecorded molecule--protein pairs are not explicitly sampled as
negative labels.

At inference, GenFish scores every candidate protein by the log-likelihood of
its stored identifier:
\begin{equation}
s_\theta(x,p)
=
\log P_\theta(\mathbf c(p)\mid x)
=
\sum_{j=1}^{7}
\log P_{\theta,j}\!\left(c_j(p)\mid x,c_{j-1}(p)\right).
\label{eq:retrieval-score}
\end{equation}
The catalogue proteins are ranked by this score and the highest-ranked
candidates are returned. Because each candidate has a known valid identifier,
its likelihood can be evaluated directly without unconstrained beam search.
These scores represent the model's relative preference over target identifiers
and are used for retrieval rather than interpreted as experimental binding
probabilities.

\section{ProTargetBench}\label{sec:benchmark}
\begin{figure}[H]
\centering
\vspace{-0.08in}\includegraphics[width=0.90\linewidth]{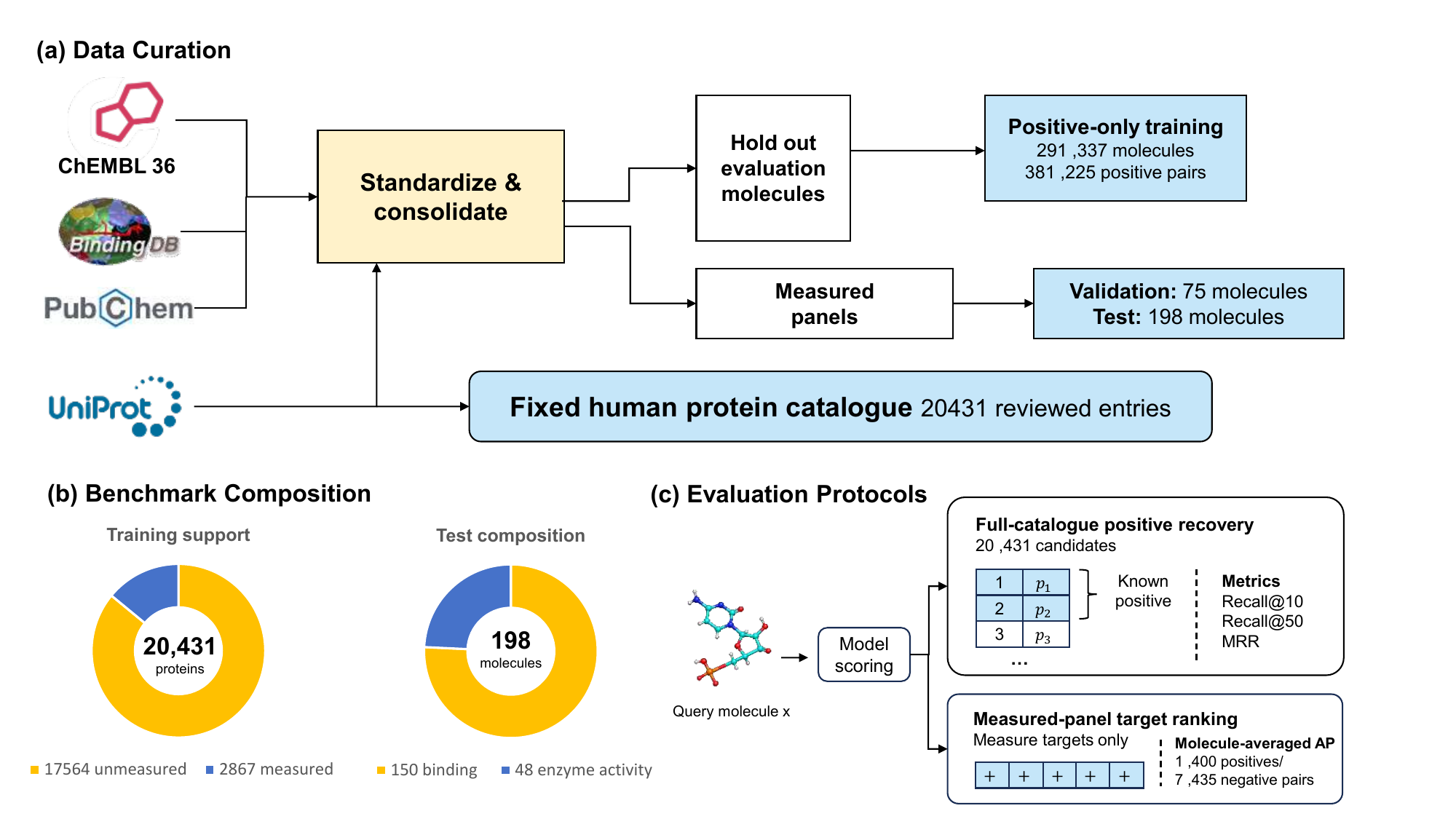}\vspace{-0.06in}
\caption{\textbf{ProTargetBench construction and evaluation.} The benchmark consolidates bioactivity evidence, fixes a human protein catalogue, and separates full-catalogue positive recovery from measured-panel target ranking.}
\label{fig:benchmark-overview}
\end{figure}
ProTargetBench separates measured-panel discrimination from recovery of known
positives in a fixed catalogue of 20,431 UniProt-reviewed human proteins,
while preserving the distinction between experimental inactivity and
unmeasured pairs.

\subsection{Benchmark Construction}

ProTargetBench integrates ChEMBL 36, BindingDB, and selected PubChem assays
\citep{chembl2019,bindingdb2025,pubchem2025}. We standardize identities and
retain eligible sub-micromolar $K_i$, $K_d$, and IC$_{50}$ positives. Evaluation
panels preserve source-specific positive and negative evidence; unmeasured
pairs are not treated as inactive.

We provide fixed training, validation, and test partitions with scaffold and
source-linked exclusions applied before aggregation. Full source versions,
identity rules, label criteria, and exclusion procedures are documented in
Appendix~\ref{app:data-contract}.

\subsection{Catalogue, Supervision, and Splits}

The catalogue contains 20,431 proteins and training has 381,225 positive
associations across 291,337 molecules. Validation and test sets contain 75 and
198 queries with 8,835 labeled test pairs.

Training-supported proteins form a subset of the catalogue, while measured test targets are defined only by held-out panels. The catalogue defines the search space; experimental records define supervision and evaluation labels. Catalogue coverage alone does not establish generalization to unsupported targets.

\subsection{Evaluation Protocols}

We evaluate the same model scores under two complementary protocols.
Full-catalogue positive recovery evaluates whether a method
can recover recorded positive targets from the entire protein catalogue.
For each query molecule $x$, all 20,431 candidates are ranked
against its recorded evaluation positives $\mathcal R_x$.
We report Recall@10, Recall@50, and mean reciprocal rank
(MRR), with
\begin{equation}
\operatorname{Recall@K}(x)
=
\frac{
|\operatorname{TopK}_\theta(x;\mathcal P)\cap\mathcal R_x|
}{
|\mathcal R_x|
}.
\end{equation}
Reciprocal rank is computed from the highest-ranked known
positive, and all retrieval metrics are averaged across query
molecules. Unmeasured candidates remain unlabeled: the protocol
assesses known-positive recovery rather than assigning
experimental inactivity to the rest of the catalogue.

Measured-panel discrimination evaluates whether a method
can distinguish active from inactive targets within the
experimentally labeled panel for each molecule. We compute
average precision (AP) for panels containing both label classes
and report molecule-averaged results separately for binding and
enzyme activity. Overall AP averages the per-molecule values
across all 198 test queries. Metric definitions, tie handling,
and aggregation details are provided in
Appendix~\ref{app:metrics}.

Both protocols reuse the same query molecules and model scores,
but assess distinct capabilities: target recovery across a broad
candidate space and activity discrimination within measured
panels. Together, the curated data, fixed splits, and evaluation
protocols provide a common basis for developing and comparing
reverse target-fishing methods without conflating missing
measurements with experimental negatives.

\section{Experiments}\label{sec:results}
We evaluate GenFish against neural and traditional baselines,
then examine false-negative competition and the biological structure
of its semantic codes. A supplementary protein--code assignment control
is provided in Appendix~\ref{app:code-assignment}.

We use the ProTargetBench training associations, 75 validation molecules, and 198 test molecules. All methods rank the same catalogue of 20,431 proteins. Comparators include four neural methods---DrugCLIP, HypSeek, LigUnity, and ConPLex---and three simple baselines: target frequency, similarity-weighted k-NN, and SEA.

The task is defined over sequence-derived protein representations. Methods whose original implementations use different protein inputs or supervision are therefore adapted to this common setting; configurations and implementation details are provided in Appendix~\ref{app:transfer}.

Headline neural results use five seeds and a ceiling of 22,560 updates. Checkpoints are selected lexicographically by validation R@10, R@50, and MRR; non-neural baselines use training or validation data only for their fixed statistics and hyperparameters. Panel evaluation reuses the catalogue-retrieval scores without reselection. Internal ablations use a common training recipe and three paired seeds. Statistical intervals are obtained by paired molecule bootstrap and are conditional on the fitted models.

\subsection{Main Results}
\begin{table}[!htb]
\centering
\footnotesize
\setlength{\tabcolsep}{3.5pt}
\caption{\textbf{Full-catalogue retrieval.} All methods rank the same 20,431-protein catalogue for 198 held-out molecules. Neural results are means $\pm$ sample SD over five seeds; non-neural baselines report single results. Higher is better ($\uparrow$).}
\label{tab:retrieval-results}
\begin{tabular}{@{}l r r r@{}}
\toprule

\multicolumn{4}{@{}l}{\emph{198 queries $\times$ 20{,}431 proteins}}\\
Method & R@10 $\uparrow$ & R@50 $\uparrow$ & MRR $\uparrow$\\
\midrule
Target frequency
& $0.0151$ & $0.1364$ & $0.0352$\\
Weighted k-NN
& $0.3483$ & $0.4187$ & $0.4227$\\
SEA (Morgan)
& $0.3074$ & $0.4092$ & $0.3534$\\
\midrule
DrugCLIP
& $0.2959\pm0.0113$ & $0.3719\pm0.0356$ & $0.4144\pm0.0131$\\
HypSeek
& $0.2598\pm0.0176$ & $0.3402\pm0.0397$ & $0.3506\pm0.0194$\\
LigUnity
& $0.3286\pm0.0104$ & $0.4126\pm0.0169$ & $0.4313\pm0.0058$\\
ConPLex
& $0.2690\pm0.0117$ & $0.4650\pm0.0226$ & $0.3147\pm0.0385$\\
GenFish
& $\mathbf{0.3527\pm0.0052}$
& $\mathbf{0.5196\pm0.0177}$
& $\mathbf{0.4631\pm0.0163}$\\
\midrule

\end{tabular}
\end{table}

\textbf{Full-catalogue retrieval.} Full-catalogue retrieval asks whether a method can recover a molecule's
recorded positive targets while ranking the entire 20,431-protein catalogue,
including candidates with no measured association for that molecule. The
results support the claim that semantic identifiers transfer sparse positive
supervision across this broad candidate space. GenFish gains are largest for
broader retrieval, where a query-independent popularity prior and local ligand
similarity provide weaker coverage. This is consistent with the intended role
of shared protein structure: improving known-target recovery without labeling
unmeasured pairs as negative.

\begin{table}[!htb]
\centering
\footnotesize
\setlength{\tabcolsep}{3.5pt}
\caption{\textbf{Measured-panel discrimination.} All methods reuse their catalogue-retrieval scores on the experimentally labeled panels. Overall AP is the molecule-averaged AP across binding and enzyme-activity queries. Higher is better ($\uparrow$).}
\label{tab:panel-results}
\begin{tabular}{@{}l r r r@{}}
\toprule
\multicolumn{4}{@{}l}{\textbf{Measured-panel discrimination}}\\
\multicolumn{4}{@{}l}{\emph{150 binding and 48 enzyme-activity molecules}}\\
Method & Binding AP $\uparrow$ & Enzyme AP $\uparrow$ & Overall AP $\uparrow$\\
\midrule

Target frequency
& $0.3433$ & $0.7212$ & $0.4349$\\
Weighted k-NN
& $0.4339$ & $0.7152$ & $0.5021$\\
SEA (Morgan)
& $0.5020$ & $0.7686$ & $0.5666$\\
\midrule
DrugCLIP
& $0.4707\pm0.0167$ & $0.7844\pm0.0209$ & $0.5468\pm0.0097$\\
HypSeek
& $0.4425\pm0.0204$ & $0.7902\pm0.0327$ & $0.5268\pm0.0091$\\
LigUnity
& $0.5000\pm0.0189$
& $\mathbf{0.8197\pm0.0255}$
& $0.5775\pm0.0146$\\
ConPLex
& $0.4442\pm0.0199$ & $0.7035\pm0.1182$ & $0.5071\pm0.0295$\\
GenFish
& $\mathbf{0.5134\pm0.0148}$
& $0.8122\pm0.0264$
& $\mathbf{0.5858\pm0.0143}$\\
\bottomrule
\end{tabular}\end{table}

\textbf{Measured-panel discrimination.} This task ranks experimentally measured candidates for each molecule. GenFish achieves the strongest overall AP and binding AP, while LigUnity remains strongest on enzyme-activity panels, showing that catalogue recovery and measured activity discrimination reward complementary evidence.

\subsection{Ablation Study}\label{sec:ablations}
\textbf{False-negative competition.}\label{sec:negative-supervision} We hide 41,290 known training positives while retaining at least one visible association for every molecule and training-supported target. The hidden labels are withheld from training but retained for identifying false-negative competition.

We compare standard contrastive training, oracle masking of hidden positives from negative competition, matched random masking, and GenFish trained on the same visible positives. All use fixed checkpoints after 22,560 updates and are evaluated on hidden-positive recovery and the original 198 held-out queries (Appendix~\ref{app:negative-supervision}).
\begin{figure}[H]
\centering
\includegraphics[width=0.72\linewidth]{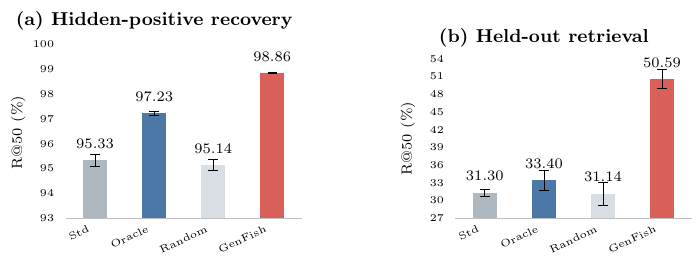}
\caption{\textbf{False-negative competition ablation.} Hidden-positive recovery (left) and held-out retrieval (right) under standard contrastive training, oracle masking, matched random masking, and GenFish with the same visible positives. Bars show means over three seeds with sample-standard-deviation error bars.}
\label{fig:negative-supervision}
\end{figure}

Oracle masking increases held-out R@50 from 31.30\% to 33.40\% and hidden-positive recovery from 95.33\% to 97.23\%, confirming that false-negative competition harms contrastive training.

Under the same incomplete supervision, GenFish reaches 50.59\% held-out R@50, exceeding standard contrastive training and oracle masking by 19.29 and 17.19 points, respectively. The intervention establishes that false-negative competition harms this contrastive model, while the comparison demonstrates GenFish's stronger target recovery under the same available positive supervision.

\subsection{Biological Structure of the Codes}\label{sec:code-semantics}

\begin{figure}[H]
\centering\vspace{-0.06in}\includegraphics[width=\linewidth]{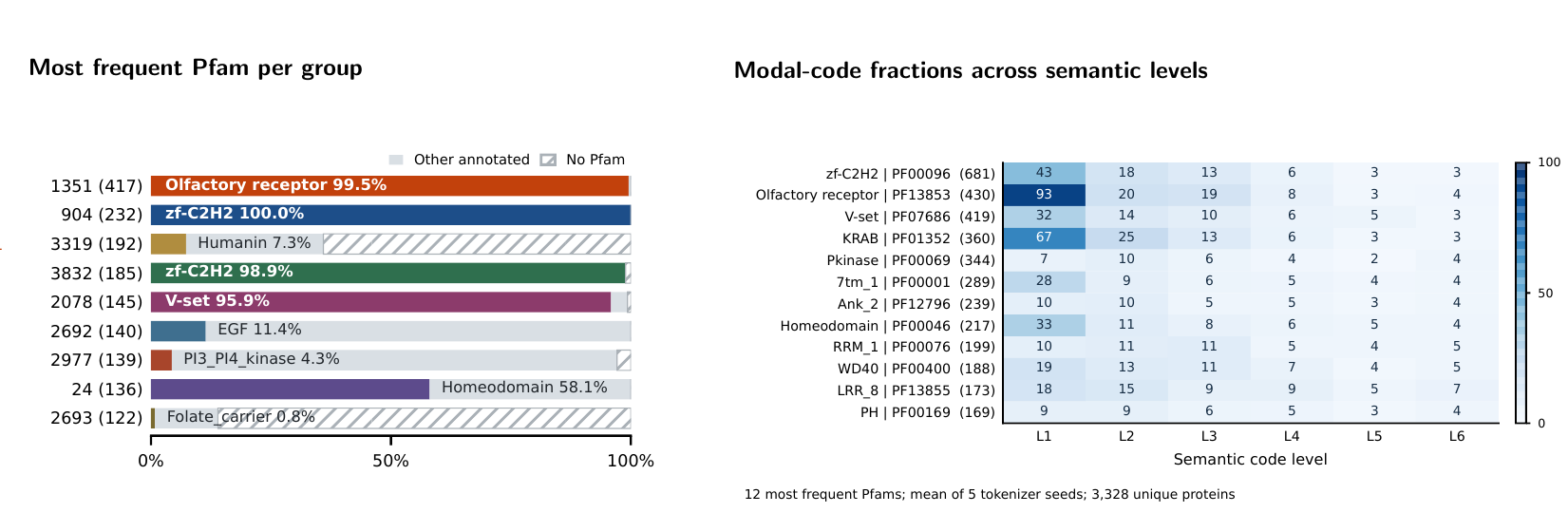}\vspace{-0.04in}
\caption{\textbf{Biological structure of the semantic identifiers.} Left: Pfam composition of large level-1 code groups. Right: modal-code fractions for prevalent Pfams across semantic levels. Pfam annotations are used only for this post hoc analysis and are excluded from tokenizer and retrieval training.}
\label{fig:code-semantics}
\end{figure}

We use Pfam annotations, held out from tokenizer and retrieval training, to
assess code--family structure. Several large groups contain coherent families;
across five tokenizer seeds, $34.98\%\pm1.34\%$ of annotated pairs sharing a
level-1 code also share a Pfam, compared with 0.47\% of all annotated catalogue-protein pairs
 (Appendix~\ref{app:code-visualization}).

Together, these results support GenFish and ProTargetBench as a framework for
retrieval under incomplete supervision. The quantized codes retain family
structure from the protein representations, suggesting that this approach can
provide a useful template for other sparse scientific retrieval problems.

\section{Conclusion and Future Work}
GenFish combines shared protein semantic identifiers with molecule-conditioned prediction for reverse target fishing under incomplete supervision. Quantized codes let observed molecule--protein associations share supervision across candidates without treating every unmeasured pair as a negative, while ProTargetBench evaluates both full-catalogue recovery and measured-panel discrimination. GenFish improves retrieval across these settings, and controlled interventions show both the cost of false-negative competition and the value of the learned protein--code structure. These results support semantic identifiers as a practical strategy for sparse proteome-scale retrieval. The framework can be extended to other molecular modalities, biological entities, and scientific discovery problems with partially observed associations.

\typeout{MAIN-CONTENT-END-PAGE: \thepage}
\subsection*{Ethics statement}
This study uses existing biochemical data, with no new human-subject or wet-laboratory experiments. Predictions are research hypotheses, not clinical recommendations or confirmed binding. Selective measurement and assay heterogeneity limit downstream use; releases must respect source licences and preserve provenance.

\subsection*{Reproducibility Statement}
The main text and appendices specify data sources,
curation rules, evaluation protocols, model architectures,
training configurations, and statistical procedures.
Appendix~\ref{app:data-contract} documents data provenance
and artifact formats; Appendices~\ref{app:tokenizer-implementation}
and~\ref{app:transfer} detail model implementations;
Appendix~\ref{app:ablations} specifies the controlled
experiments and internal ablations.

\bibliography{references}

\begin{thebibliography}{24}
\providecommand{\natexlab}[1]{#1}
\providecommand{\url}[1]{\texttt{#1}}
\expandafter\ifx\csname urlstyle\endcsname\relax
  \providecommand{\doi}[1]{doi: #1}\else
  \providecommand{\doi}{doi: \begingroup \urlstyle{rm}\Url}\fi

\bibitem[B{\'e}quignon et~al.(2023)B{\'e}quignon, Bongers, Jespers, IJzerman,
  van~der Water, and van Westen]{papyrus2023}
O.~J.~M. B{\'e}quignon, B.~J. Bongers, W.~Jespers, A.~P. IJzerman, B.~van~der
  Water, and G.~J.~P. van Westen.
\newblock {Papyrus}: a large-scale curated dataset aimed at bioactivity
  predictions.
\newblock \emph{Journal of Cheminformatics}, 15\penalty0 (1):\penalty0 3, 2023.
\newblock \doi{10.1186/s13321-022-00672-x}.

\bibitem[Candido et~al.(2026)Candido, Hayes, Derry, Rao, Lin, Verkuil, Wu, Lee,
  Bruguera, Keval, Kopylov, Pak, Wu, Thomas, Mataraso, Hsu, Trotman-Grant,
  Fatras, dos Santos~Costa, Badkundri, Ak{\i}n, Oktay, Deaton, Montabana,
  Sitwala, Yu, Wiggert, Carlin, Goering, Blazejewski, Sandora, Hla, Jia,
  Kloker, Sofroniew, Uehara, Pannu, Bachas, Liu, Sercu, and Rives]{esmc}
Salvatore Candido, Thomas Hayes, Alexander Derry, Roshan Rao, Zeming Lin,
  Robert Verkuil, Bryan Wu, Jin~Sub Lee, Elise~S. Bruguera, Jehan~A. Keval,
  Mykhailo Kopylov, John~E. Pak, Wesley Wu, Neil Thomas, Samson Mataraso, Alvin
  Hsu, Ashton~C. Trotman-Grant, Kilian Fatras, Allan dos Santos~Costa, Rohil
  Badkundri, Halil Ak{\i}n, Deniz Oktay, Jonathan Deaton, Elizabeth Montabana,
  Hrishita Sitwala, Yue Yu, Marius Wiggert, Dylan~Alexander Carlin, Anthony~W.
  Goering, Tomasz Blazejewski, McCullen Sandora, Michael Hla, Tina~Z. Jia,
  Leon~H. Kloker, Nicholas~J. Sofroniew, Masatoshi Uehara, Jassi Pannu, Sharrol
  Bachas, Daniel~S. Liu, Tom Sercu, and Alexander Rives.
\newblock Language modeling materializes a world model of protein biology,
  2026.
\newblock URL \url{https://www.biorxiv.org/content/10.64898/2026.06.03.729735}.
\newblock bioRxiv preprint.

\bibitem[Daina et~al.(2019)Daina, Michielin, and Zoete]{swisstarget2019}
Antoine Daina, Olivier Michielin, and Vincent Zoete.
\newblock {SwissTargetPrediction}: updated data and new features for efficient
  prediction of protein targets of small molecules.
\newblock \emph{Nucleic Acids Research}, 47\penalty0 (W1):\penalty0 W357--W364,
  2019.
\newblock \doi{10.1093/nar/gkz382}.
\newblock URL \url{https://doi.org/10.1093/nar/gkz382}.

\bibitem[Feng et~al.(2025)Feng, Liu, Li, Yang, Zou, Cao, Li, Zhang, and
  Wang]{ligunity2025}
Bin Feng, Zijing Liu, Hao Li, Mingjun Yang, Junjie Zou, He~Cao, Yu~Li, Lei
  Zhang, and Sheng Wang.
\newblock Hierarchical affinity landscape navigation through learning a shared
  pocket-ligand space.
\newblock \emph{Patterns}, 2025.
\newblock \doi{10.1016/j.patter.2025.101371}.

\bibitem[Gao et~al.(2023)Gao, Qiang, Tan, Jia, Ren, Lu, Liu, Ma, and
  Lan]{drugclip2023}
Bowen Gao, Bo~Qiang, Haichuan Tan, Yinjun Jia, Minsi Ren, Minsi Lu, Jingjing
  Liu, Wei-Ying Ma, and Yanyan Lan.
\newblock {DrugCLIP}: Contrastive protein-molecule representation learning for
  virtual screening.
\newblock In \emph{Advances in Neural Information Processing Systems},
  volume~36, pages 44595--44614, 2023.
\newblock \doi{10.52202/075280-1932}.
\newblock URL \url{https://doi.org/10.52202/075280-1932}.

\bibitem[Huang et~al.(2021)Huang, Xiao, Glass, and Sun]{moltrans2021}
Kexin Huang, Cao Xiao, Lucas~M. Glass, and Jimeng Sun.
\newblock {MolTrans}: Molecular interaction transformer for drug--target
  interaction prediction.
\newblock \emph{Bioinformatics}, 37\penalty0 (6):\penalty0 830--836, 2021.
\newblock \doi{10.1093/bioinformatics/btaa880}.
\newblock URL \url{https://doi.org/10.1093/bioinformatics/btaa880}.

\bibitem[Keiser et~al.(2007)Keiser, Roth, Armbruster, Ernsberger, Irwin, and
  Shoichet]{sea2007}
Michael~J. Keiser, Bryan~L. Roth, Blaine~N. Armbruster, Paul Ernsberger,
  John~J. Irwin, and Brian~K. Shoichet.
\newblock Relating protein pharmacology by ligand chemistry.
\newblock \emph{Nature Biotechnology}, 25\penalty0 (2):\penalty0 197--206,
  2007.
\newblock \doi{10.1038/nbt1284}.
\newblock URL \url{https://doi.org/10.1038/nbt1284}.

\bibitem[Keiser et~al.(2009)Keiser, Setola, Irwin, Laggner, Abbas, Hufeisen,
  Jensen, Kuijer, Matos, Tran, Whaley, Glennon, Hert, Thomas, Edwards,
  Shoichet, and Roth]{keiser2009}
Michael~J. Keiser, Vincent Setola, John~J. Irwin, Christian Laggner, Atheir~I.
  Abbas, Sandra~J. Hufeisen, Niels~H. Jensen, Michael~B. Kuijer, Roberto~C.
  Matos, Thuy~B. Tran, Ryan Whaley, Richard~A. Glennon, J{\'e}r{\^o}me Hert,
  Kelan L.~H. Thomas, Douglas~D. Edwards, Brian~K. Shoichet, and Bryan~L. Roth.
\newblock Predicting new molecular targets for known drugs.
\newblock \emph{Nature}, 462\penalty0 (7270):\penalty0 175--181, 2009.
\newblock \doi{10.1038/nature08506}.
\newblock URL \url{https://doi.org/10.1038/nature08506}.

\bibitem[Kim et~al.(2025)Kim, Chen, Cheng, Gindulyte, He, He, Li, Shoemaker,
  Thiessen, Yu, Zaslavsky, Zhang, and Bolton]{pubchem2025}
Sunghwan Kim, Jie Chen, Tiejun Cheng, Asta Gindulyte, Jia He, Siqian He,
  Qingliang Li, Benjamin~A. Shoemaker, Paul~A. Thiessen, Bo~Yu, Leonid
  Zaslavsky, Jian Zhang, and Evan~E. Bolton.
\newblock {PubChem} 2025 update.
\newblock \emph{Nucleic Acids Research}, 53\penalty0 (D1):\penalty0
  D1516--D1525, 2025.
\newblock \doi{10.1093/nar/gkae1059}.
\newblock URL \url{https://doi.org/10.1093/nar/gkae1059}.

\bibitem[Li et~al.(2006)Li, Gao, Kang, Zhang, Yang, Yu, Luo, Zhu, Chen, Shen,
  Wang, and Jiang]{tarfisdock2006}
Honglin Li, Zhenting Gao, Ling Kang, Hailei Zhang, Kun Yang, Kunqian Yu,
  Xiaomin Luo, Weiliang Zhu, Kaixian Chen, Jianhua Shen, Xicheng Wang, and
  Hualiang Jiang.
\newblock {TarFisDock}: a web server for identifying drug targets with docking
  approach.
\newblock \emph{Nucleic Acids Research}, 34\penalty0 (Web Server
  issue):\penalty0 W219--W224, 2006.
\newblock \doi{10.1093/nar/gkl114}.
\newblock URL \url{https://doi.org/10.1093/nar/gkl114}.

\bibitem[Li et~al.(2025)Li, Ren, Wang, Han, Li, Li, and
  Bai]{maitargetfisher2025}
Shi-Wei Li, Peng-Xuan Ren, Lin Wang, Qi-Lei Han, Feng-Lei Li, Hong-Lin Li, and
  Fang Bai.
\newblock {MAI-TargetFisher}: A proteome-wide drug target prediction method
  synergetically enhanced by artificial intelligence and physical modeling.
\newblock \emph{Acta Pharmacologica Sinica}, 46\penalty0 (5):\penalty0
  1462--1475, 2025.
\newblock \doi{10.1038/s41401-024-01444-z}.
\newblock URL \url{https://doi.org/10.1038/s41401-024-01444-z}.

\bibitem[Liu et~al.(2025)Liu, Hwang, Burley, Nitsche, Southan, Walters, and
  Gilson]{bindingdb2025}
Tiqing Liu, Linda Hwang, Stephen~K. Burley, Carmen~I. Nitsche, Christopher
  Southan, W.~Patrick Walters, and Michael~K. Gilson.
\newblock {BindingDB} in 2024: a {FAIR} knowledgebase of protein-small molecule
  binding data.
\newblock \emph{Nucleic Acids Research}, 53\penalty0 (D1):\penalty0
  D1633--D1644, 2025.
\newblock \doi{10.1093/nar/gkae1075}.
\newblock URL \url{https://doi.org/10.1093/nar/gkae1075}.

\bibitem[Luo et~al.(2024)Luo, Wang, Li, Zheng, Mu, and Guo]{luo2024reverse}
Qing Luo, Sheng Wang, Hoi~Yeung Li, Liangzhen Zheng, Yuguang Mu, and Jingjing
  Guo.
\newblock Benchmarking reverse docking through {AlphaFold2} human proteome.
\newblock \emph{Protein Science}, 33\penalty0 (10):\penalty0 e5167, 2024.
\newblock \doi{10.1002/pro.5167}.

\bibitem[Mayr et~al.(2018)Mayr, Klambauer, Unterthiner, Steijaert, Wegner,
  Ceulemans, Clevert, and Hochreiter]{mayr2018}
Andreas Mayr, G{\"u}nter Klambauer, Thomas Unterthiner, Marvin Steijaert,
  J{\"o}rg~K. Wegner, Hugo Ceulemans, Djork-Arn{\'e} Clevert, and Sepp
  Hochreiter.
\newblock Large-scale comparison of machine learning methods for drug target
  prediction on {ChEMBL}.
\newblock \emph{Chemical Science}, 9\penalty0 (24):\penalty0 5441--5451, 2018.
\newblock \doi{10.1039/C8SC00148K}.
\newblock URL \url{https://doi.org/10.1039/C8SC00148K}.

\bibitem[Mendez et~al.(2019)Mendez, Gaulton, Bento, Chambers, De~Veij,
  F{\'e}lix, Magari{\~n}os, Mosquera, Mutowo, Nowotka,
  Gordillo-Mara{\~n}{\'o}n, Hunter, Junco, Mugumbate, Rodriguez-Lopez,
  Atkinson, Bosc, Radoux, Segura-Cabrera, Hersey, and Leach]{chembl2019}
David Mendez, Anna Gaulton, A.~Patr{\'i}cia Bento, Jon Chambers, Marleen
  De~Veij, Eloy F{\'e}lix, Mar{\'i}a~Paula Magari{\~n}os, Juan~F. Mosquera,
  Prudence Mutowo, Micha{\l} Nowotka, Mar{\'i}a Gordillo-Mara{\~n}{\'o}n, Fiona
  Hunter, Laura Junco, Grace Mugumbate, Milagros Rodriguez-Lopez, Francis
  Atkinson, Nicolas Bosc, Chris~J. Radoux, Aldo Segura-Cabrera, Anne Hersey,
  and Andrew~R. Leach.
\newblock {ChEMBL}: towards direct deposition of bioassay data.
\newblock \emph{Nucleic Acids Research}, 47\penalty0 (D1):\penalty0 D930--D940,
  2019.
\newblock \doi{10.1093/nar/gky1075}.
\newblock URL \url{https://doi.org/10.1093/nar/gky1075}.

\bibitem[Nguyen et~al.(2021)Nguyen, Le, Quinn, Nguyen, Le, and
  Venkatesh]{graphdta2021}
Thin Nguyen, Hang Le, Thomas~P. Quinn, Tri Nguyen, Thuc~Duy Le, and Svetha
  Venkatesh.
\newblock {GraphDTA}: predicting drug--target binding affinity with graph
  neural networks.
\newblock \emph{Bioinformatics}, 37\penalty0 (8):\penalty0 1140--1147, 2021.
\newblock \doi{10.1093/bioinformatics/btaa921}.
\newblock URL \url{https://doi.org/10.1093/bioinformatics/btaa921}.

\bibitem[{\"O}zt{\"u}rk et~al.(2018){\"O}zt{\"u}rk, {\"O}zg{\"u}r, and
  Ozkirimli]{deepdta2018}
Hakime {\"O}zt{\"u}rk, Arzucan {\"O}zg{\"u}r, and Elif Ozkirimli.
\newblock {DeepDTA}: deep drug--target binding affinity prediction.
\newblock \emph{Bioinformatics}, 34\penalty0 (17):\penalty0 i821--i829, 2018.
\newblock \doi{10.1093/bioinformatics/bty593}.
\newblock URL \url{https://doi.org/10.1093/bioinformatics/bty593}.

\bibitem[Pe{\'o}n et~al.(2016)Pe{\'o}n, Dang, and Ballester]{peon2016}
Antonio Pe{\'o}n, Cuong~C Dang, and Pedro~J Ballester.
\newblock How reliable are ligand-centric methods for target fishing?
\newblock \emph{Frontiers in Chemistry}, 4:\penalty0 15, 2016.
\newblock \doi{10.3389/fchem.2016.00015}.
\newblock URL \url{https://doi.org/10.3389/fchem.2016.00015}.

\bibitem[Rajput et~al.(2023)Rajput, Mehta, Singh, Keshavan, Vu, Heldt, Hong,
  Tay, Tran, Samost, Kula, Chi, and Sathiamoorthy]{tiger2023}
Shashank Rajput, Nikhil Mehta, Anima Singh, Raghunandan~Hulikal Keshavan, Trung
  Vu, Lukasz Heldt, Lichan Hong, Yi~Tay, Vinh Tran, Jonah Samost, Maciej Kula,
  Ed~Chi, and Maheswaran Sathiamoorthy.
\newblock Recommender systems with generative retrieval.
\newblock In \emph{Advances in Neural Information Processing Systems},
  volume~36, pages 10299--10315, 2023.
\newblock \doi{10.52202/075280-0452}.
\newblock URL \url{https://arxiv.org/abs/2305.05065}.

\bibitem[Singh et~al.(2023)Singh, Sledzieski, Bryson, Cowen, and
  Berger]{conplex2023}
Rohit Singh, Samuel Sledzieski, Bryan Bryson, Lenore Cowen, and Bonnie Berger.
\newblock Contrastive learning in protein language space predicts interactions
  between drugs and protein targets.
\newblock \emph{Proceedings of the National Academy of Sciences}, 120\penalty0
  (24), 2023.
\newblock \doi{10.1073/pnas.2220778120}.
\newblock URL \url{https://doi.org/10.1073/pnas.2220778120}.

\bibitem[Tay et~al.(2022)Tay, Tran, Dehghani, Ni, Bahri, Mehta, Qin, Hui, Zhao,
  Gupta, Schuster, Cohen, and Metzler]{dsi2022}
Yi~Tay, Vinh~Q. Tran, Mostafa Dehghani, Jianmo Ni, Dara Bahri, Harsh Mehta,
  Zhen Qin, Kai Hui, Zhe Zhao, Jai Gupta, Tal Schuster, William~W. Cohen, and
  Donald Metzler.
\newblock Transformer memory as a differentiable search index.
\newblock In \emph{Advances in Neural Information Processing Systems},
  volume~35, pages 21831--21843, 2022.
\newblock \doi{10.52202/068431-1587}.
\newblock URL \url{https://arxiv.org/abs/2202.06991}.

\bibitem[Wang et~al.(2026)Wang, Zhu, Gao, Hong, Zhang, Ma, and
  Lan]{hypseek2025}
Jianhui Wang, Wenyu Zhu, Bowen Gao, Xin Hong, Ya-Qin Zhang, Wei-Ying Ma, and
  Yanyan Lan.
\newblock Learning protein--ligand binding in hyperbolic space.
\newblock In \emph{Proceedings of the AAAI Conference on Artificial
  Intelligence}, volume~40, pages 1150--1158, 2026.
\newblock \doi{10.1609/aaai.v40i2.37086}.
\newblock URL \url{https://doi.org/10.1609/aaai.v40i2.37086}.

\bibitem[Wang et~al.(2024)Wang, Bao, Lin, Zhang, Li, Feng, Ng, and
  Chua]{letter2024}
Wenjie Wang, Honghui Bao, Xinyu Lin, Jizhi Zhang, Yongqi Li, Fuli Feng,
  See-Kiong Ng, and Tat-Seng Chua.
\newblock Learnable item tokenization for generative recommendation.
\newblock In \emph{Proceedings of the 33rd ACM International Conference on
  Information and Knowledge Management}, pages 2400--2409, 2024.
\newblock \doi{10.1145/3627673.3679569}.
\newblock URL \url{https://doi.org/10.1145/3627673.3679569}.

\bibitem[Zong et~al.(2022)Zong, Li, Wen, Ngo, Yu, Huang, Chowdhury, Jiang, Fu,
  Weinshilboum, Jiang, Hunter, and Liu]{beta2022}
Nansu Zong, Ning Li, Andrew Wen, Victoria Ngo, Yue Yu, Ming Huang, Shaika
  Chowdhury, Chao Jiang, Sunyang Fu, Richard Weinshilboum, Guoqian Jiang,
  Lawrence Hunter, and Hongfang Liu.
\newblock {BETA}: a comprehensive benchmark for computational drug--target
  prediction.
\newblock \emph{Briefings in Bioinformatics}, 23\penalty0 (4):\penalty0
  bbac199, 2022.
\newblock \doi{10.1093/bib/bbac199}.
\newblock URL \url{https://doi.org/10.1093/bib/bbac199}.

\end{thebibliography}
\bibliographystyle{plainnat}
\clearpage
\appendix

The appendices provide benchmark construction and
evaluation details
(Appendix~\ref{app:data-contract}),
model implementation
(Appendix~\ref{app:tokenizer-implementation}),
probability and scoring identities
(Appendix~\ref{app:math}),
external comparator configurations
(Appendix~\ref{app:transfer}),
and controlled experiments and internal ablations
(Appendix~\ref{app:ablations}).

\section{Benchmark Construction, Splits, and Metrics}
\label{app:data-contract}

This appendix describes the data sources, curation rules,
evaluation splits, metrics, and release contents of
ProTargetBench.

\subsection{Sources and Benchmark Overview}

ProTargetBench combines a fixed human protein catalogue
with curated bioactivity associations and experimentally
measured evaluation panels. Construction standardizes
molecular and protein identities, assesses measurement
eligibility, applies evaluation exclusions, and aggregates
the remaining evidence into training associations.

The v1 data snapshot (2026-09-10) uses ChEMBL~36,
BindingDB All~202609 with assay and record sidecars, and
selected reviewed PubChem panels. The PubChem component
is a panel collection rather than a database-wide extraction.
The human catalogue uses UniProt release 2026\_03 with
the query
\texttt{organism\_id:9606 AND reviewed:true}.

The catalogue contains 20,431 protein accessions and
20,353 distinct sequences. Distinct accessions are retained
even when their sequences are identical; entries therefore
do not correspond strictly to distinct genes. Of the
catalogue entries, 2,303 exceed 1,022 residues and 17,564
have no positive training support.
The benchmark overview in Section~\ref{sec:benchmark} summarizes the main
benchmark counts.

The feature store contains 20,685 protein identities,
including identities outside the retrieval catalogue.
These additional identities do not expand the candidate
set. Full-catalogue predictions follow a fixed ordering
of the 20,431 candidate accessions.

\subsection{Identity Standardization and Evidence Curation}

\paragraph{Molecular identities.}
Molecules are standardized with RDKit~2023.09.6.
The pipeline parses SMILES, rejects unparseable structures
and enhanced-stereochemistry groups, removes atom maps,
and applies RDKit Cleanup. Registered salts are stripped
with \texttt{dontRemoveEverything=True}; unresolved
multicomponent structures and nonorganic parents are
excluded, and the retained parent is uncharged.

Canonical isomeric SMILES preserves stereochemistry and
isotopes; tautomers are not collapsed. Molecular identifiers
are SHA256 hashes of these canonical strings.
Nonchiral Bemis--Murcko scaffolds are hashed separately
for exclusion checks; molecules with empty scaffolds have
no scaffold key. The use of nonchiral Morgan fingerprints
as model inputs does not change the molecular identity
policy.

\paragraph{Protein identities.}
Association-level protein identities are defined by
the tuple of taxon, accession, and sequence SHA256,
rather than accession alone. Primary training evidence
requires resolved taxonomy, an admissible sequence,
a single protein component, and a canonical construct.
Sequences that conflict with the catalogue reference
are not silently replaced.

Historical evaluation-panel candidates retain their
frozen sequence identities when an exact canonical match
is unavailable. The retrieval catalogue retains distinct
accessions rather than merging them by sequence.

\paragraph{Activity measurements and training labels.}
Eligible Ki, Kd, and IC50 measurements are converted
to nM using decimal arithmetic. Supported relations are
$=,<,\leq,>,\geq$, with open and closed interval endpoints
preserved. Nonpositive, nonfinite, malformed, and
ambiguously censored range records are excluded.

A measurement provides positive evidence only if its
entire admissible interval lies below 1,000\,nM.
Thus, ``$<1000$'' qualifies, whereas ``$\leq1000$''
does not. A lower bound of at least 10,000\,nM provides
weak-activity evidence, but is admitted as a credible
negative only with verified fitted-value,
experimental-bound, or validated-panel provenance.
Intermediate measurements and threshold-crossing
intervals remain nonbinary.

The training view used in this study contains positive
associations and no admitted credible negatives.
Unrecorded pairs used as competitors by a training
objective are not experimentally measured negatives.

ChEMBL training records additionally require assay
type B, confidence score 9, standard flag 1, no validity
comment or potential-duplicate flag, and no variant
sequence. Reviewed experimental contexts admit binding
or enzyme activity and exclude unresolved mixed,
phenotypic, or variant contexts. EC50 records may be
retained in the raw evidence ledger but are not admitted
as training Ki/Kd/IC50 evidence.

\paragraph{Duplicates and conflicts.}
Measurements are merged across sources only when they
share the original assay identity, standardized molecule,
protein identity, endpoint, experimental context, and
interval bounds. Otherwise, they remain source-local
records.

Within a common endpoint and context, positive evidence
combined with weak or credible-negative evidence is
flagged as conflicting. Strong measurements that conflict
with inactive-text assertions are also flagged rather
than averaged. Incomparable endpoints are not combined
into a single affinity value.

Evaluation exclusions are applied before evidence is
reaggregated. Admitted molecule--protein identity pairs
are then deduplicated so that repeated measurements
do not multiply their contribution to the training loss.

\subsection{Evaluation Panels and Split Construction}
\label{app:benchmark-details}

\paragraph{Panel sources and labels.}
Evaluation queries retain their frozen source-specific
experimental labels. Binding and enzyme panels are
evaluated separately.
\space The resource descriptions \citep{chembl2019,pubchem2025,bindingdb2025} document the databases; they do not replace the original assay evidence supporting each retained label.

Panel labels follow source-specific criteria rather
than a universal threshold. Positive labels generally
require eligible sub-micromolar measurements with
admissible primary evidence. Negative labels require
source-supported evidence, such as an explicit lower
bound of at least $10\,\mu$M or a documented inactive
call satisfying the applicable counter-screen criteria.
Conflicts between strong activity measurements and
primary inactive calls are excluded rather than averaged.

Dose-response sources require complete dose series
and admissible curve classes. Exact assay identifiers,
curve-class and response cutoffs, counter-screen rules,
and reviewed BindingDB or patent-table references are
specified in the release manifest. These source-specific
predicates are required to reproduce the panel labels;
the general thresholds above do not replace them.

\paragraph{Frozen evaluation split.}
The evaluation split, referred to as \emph{Anchor}
in the data artifacts, contains 75 validation molecules
and 198 test molecules. The validation set comprises
53 binding and 22 enzyme queries; the test set comprises
150 binding and 48 enzyme queries.

The validation and test query payloads retain their
frozen identities, labels, and source provenance.
Catalogue size, the number of training-supported
proteins, and the number of experimentally measured
panel targets describe different aspects of the
benchmark and are reported separately.

\paragraph{Training exclusions.}
Before constructing the final training view, the builder
excludes evaluation molecules, their nonempty scaffold
keys, the frozen held-out scaffold list, and historical
molecule and pair exclusion lists. Evaluated
DOI, PMID, and assay aliases are expanded through the
source-identity graph so that reviewed evidence
exclusions apply across linked sources and species.
Previously identified conflicting pairs remain flagged.

These operations filter the training evidence while
preserving the frozen validation and test queries.
They should not be interpreted as establishing additional
pairwise scaffold-disjointness guarantees between
validation and test beyond the recorded split policy.

\paragraph{Validation-based selection.}
Hyperparameters and model configurations are selected
using validation data. Learning-rate selection uses
three-seed mean validation R@10, with ties broken by
R@50 and then MRR. Both test protocols use the resulting
catalogue-selected checkpoints; measured-panel test AP
is not used for checkpoint reselection.

Derived panels retain their parent-source provenance
and are not treated as newly collected external data.

\subsection{Metrics and Statistical Reporting}
\label{app:metrics}

\paragraph{Full-catalogue retrieval.}
Let $\mathcal{P}_x^{+}$ denote the set of recorded
positive catalogue targets for query molecule $x$,
and let $\operatorname{TopK}(x)$ denote the first $K$
proteins in its ranked catalogue. Recall is
\begin{equation}
\operatorname{R@K}(x)
=
\frac{
|\mathcal{P}_x^{+}\cap\operatorname{TopK}(x)|
}{
|\mathcal{P}_x^{+}|
}.
\end{equation}

The reciprocal rank is determined by the highest-ranked
recorded positive:
\begin{equation}
\operatorname{RR}(x)
=
\frac{1}{
\min_{p\in\mathcal{P}_x^{+}}
\operatorname{rank}_x(p)
}.
\end{equation}

We report molecule-macro R@10, R@50, and MRR by averaging
the corresponding query-level values over the evaluation
molecules. These metrics assess the recovery of recorded
positives; they do not establish that higher-ranked
unmeasured associations are biologically inactive.

\paragraph{Measured-panel discrimination.}
Average precision is calculated using groups of equal
prediction scores:
\begin{equation}
\operatorname{AP}(x)
=
\sum_t
\left(R_t-R_{t-1}\right)
\operatorname{Precision}_t,
\end{equation}
where $t$ indexes successive score thresholds and
$R_t$ is recall among the panel's experimentally labeled
positives. Processing tied scores as a group avoids
arbitrary ordering within ties.

Where AUROC is computed, it is the fraction of
positive--negative comparisons won by the positive,
with half credit for a tie. Scores are averaged over
queries within a molecule when needed, and then over
molecules.

We report binding-tier and enzyme-tier macro AP
separately. Overall AP is
\begin{equation}
\operatorname{AP}_{\mathrm{overall}}
=
\frac{
150\,\operatorname{AP}_{\mathrm{binding}}
+
48\,\operatorname{AP}_{\mathrm{enzyme}}
}{198}.
\end{equation}
Because the tiers are disjoint, this equals the mean
of the 198 molecule-level AP values. It is neither
AP computed from pooled predictions nor an average
weighted by panel size.

For neural methods, Overall AP is computed separately
for each seed before calculating its mean and sample
standard deviation, preserving covariance between the
two tiers. This descriptive aggregate was added after
inspection of tier-specific results, without changing
model fitting or checkpoint selection.

\paragraph{Seeds and uncertainty.}
The initial catalogue experiments use seeds 13, 37,
and 73. Seeds 101 and 211 are added as confirmatory
runs for the headline comparison, using fresh supervised
initializations and the same model recipes. For GenFish,
the tokenizer is also fitted afresh under the same
500-epoch feature-reconstruction protocol.
No replication seed is omitted.

Neural headline results report five-seed means and
sample standard deviations, with denominator $n-1$.
Internal controlled-study intervals use 10,000 paired
molecule-bootstrap resamples. The low-similarity
comparisons are specified in
Appendix~\ref{app:hard-band}. Bootstrap intervals
quantify query-sampling uncertainty conditional on
the fitted models, rather than the full uncertainty
from training or experimental-source sampling.

\subsection{Release Contents and Provenance}

Versioned manifests, policies, verification logs,
and source/hash indices identify the data views
and evaluation splits. The staged derived release
contains catalogue sequences, standardized molecular
structures, training evidence groups, query labels,
exclusion policies, a loader, and initial predictions.
It also includes historical material outside the
current evaluation. 

The loader verifies molecular structure hashes under
the identity policy above. Protein identities follow
the taxon/accession/sequence-hash policy, while
catalogue file order defines the canonical candidate
ordering and prediction records identify candidates
by UniProt accession.

The staged Anchor prediction files contain scored
top-200 lists and the full-catalogue rank of every
recorded positive, allowing exact recomputation of
R@10, R@50, and MRR. All twelve staged method/seed
files reproduce their originating metrics within
numerical precision, but do not cover every manuscript
result.

Raw source-ledger measurements are not redistributed.
Retained \texttt{evidence\_id} and
\texttt{source\_molecule\_id} fields support label-level
audit against ChEMBL~36 and BindingDB All~202609.
The derived release supports inspection of the frozen
benchmark views but does not, by itself, establish
independent reconstruction of the primary-source
splits from scratch.

\section{Model Implementation and Training}
\label{app:tokenizer-implementation}

\subsection{Protein Features and Sequence Pooling}
\label{app:sequence-pooling}

We extract frozen 1,152-dimensional protein features
using ESMC-600M \citep{esmc}. Sequences are divided into
nonoverlapping windows of at most 1,022 residues.
Residue representations exclude special tokens and
are averaged across the complete sequence.

For a sequence of length $n_p$, let
$B_p=\lceil n_p/1022\rceil$ be the number of windows,
$m_{p,b}$ the length of window $b$, and $H_{p,b,t}$
the representation of its $t$th residue. The pooled
feature is
\begin{equation}
\widetilde h_p
=
\sum_{b=1}^{B_p}
\frac{m_{p,b}}{n_p}
\left(
\frac{1}{m_{p,b}}
\sum_{t=1}^{m_{p,b}} H_{p,b,t}
\right).
\end{equation}
This gives equal weight to every residue while covering
the full sequence. Residues are encoded within their
respective windows, without cross-window context.

The pooled features are standardized coordinate-wise
using the fitting population's means and sample standard
deviations, with standard deviations clamped below at
$10^{-6}$. We denote the standardized feature by $h_p$.
Feature standardization and tokenizer fitting use all
20,685 core protein identities, including the 20,431
catalogue entries. Additional feature-store identities
are not retrieval candidates.

\subsection{Tokenizer Construction and Training}

\paragraph{Residual quantization.}
An encoder $E$ maps $h_p$ to a latent representation
$z_p\in\mathbb{R}^{32}$. At level $\ell$, a codebook
$\{e_{\ell,k}\}_{k=1}^{K_\ell}$ quantizes the current
residual:
\begin{equation}
r_0=z_p,\qquad
c_\ell(p)=
\arg\min_k
\|r_{\ell-1}-e_{\ell,k}\|_2^2,\qquad
q_\ell=e_{\ell,c_\ell(p)}.
\end{equation}
Nearest-vector ties are resolved by the lowest code
index. The six codebook sizes are
$(4096,32,128,128,256,128)$.

Each selected vector is subtracted from the residual.
The sum of selected vectors is passed through a decoder
to reconstruct the standardized protein feature.
The tokenizer uses hidden width 256 and latent
dimension 32.

\paragraph{Objective and gradient estimator.}
The straight-through estimator and reconstruction are
\begin{equation}
\widetilde q_\ell
=
r_{\ell-1}+\sg(q_\ell-r_{\ell-1}),
\qquad
r_\ell=r_{\ell-1}-\widetilde q_\ell,
\qquad
\widehat h_p
=
D\left(\sum_{\ell=1}^{6}\widetilde q_\ell\right).
\end{equation}

The training objective is
\begin{equation}
\mathcal{L}_{\mathrm{tok}}
=
\mse(\widehat h_p,h_p)
+
\sum_{\ell=1}^{6}
\left[
\mse(\sg(q_\ell),r_{\ell-1})
+
0.25\,\mse(q_\ell,\sg(r_{\ell-1}))
\right],
\end{equation}
where $\sg$ stops gradients and $\mse$ averages squared
elementwise errors. The tokenizer is trained for
500 epochs using Adam with learning rate $10^{-3}$,
then frozen for molecular predictor training.
Its objective uses protein features rather than
interaction labels.

For the displayed estimator, treating discrete
assignments as fixed,
\begin{equation}
\frac{\partial\widetilde q_\ell}
{\partial r_{\ell-1}}=I,
\qquad
\frac{\partial r_\ell}
{\partial r_{\ell-1}}=0.
\end{equation}
Consequently, later levels do not propagate gradients
to earlier levels through the residual path.
Reconstruction gradients reach the encoder through
the first straight-through path, while the explicit
codebook term updates the code vectors.

\paragraph{Collision-resolving suffix.}
Proteins are grouped by their complete six-code
semantic tuple. Within each group, consecutive suffix
integers starting at zero are assigned in the fixed
protein-feature row order. The suffix vocabulary size
equals the largest collision-group size; singleton
groups receive suffix zero.

The suffix has no reconstruction vector. It is
predicted by an independent head and participates
in both code cross-entropy and retrieval scoring.
For the internal experiments with seeds 13, 37,
and 73, its vocabulary sizes are 11, 10, and 8,
respectively. This assignment is fixed for the
catalogue; extending the catalogue may require
updating the suffix assignment and prediction
vocabulary.

\subsection{Molecular Predictor and Conditional Interaction}
\label{app:model-contract}

\paragraph{Shared predictor.}
Molecules are represented by 2,048-bit Morgan
fingerprints with radius two. A shared molecular
network uses two 512-wide ReLU layers with dropout
0.1, followed by seven code-prediction heads.
The first semantic code and the suffix are predicted
without predecessor conditioning. Five interaction
modules modify the logits at semantic levels 2--6.

\paragraph{Conditional modules.}
Each module uses interaction width 16.
For a fixed query $x$, the matrix of logit
corrections across predecessor and current codes is
\begin{equation}
\Delta_j(x)
=
\bar U_j
\operatorname{diag}(g_j(x))
V_j^\top,
\qquad
\operatorname{rank}(\Delta_j(x))\leq16.
\end{equation}
Here $\bar U_j$ contains row-normalized predecessor
embeddings, with norms clamped below at $10^{-12}$.
The rank bound applies to the logit correction,
not necessarily to the resulting softmax
probability matrix.

The five modules add
\begin{equation}
\sum_{j=2}^{6}
\left[
16(K_{j-1}+K_j)+16\cdot512+16
\right]
=
126{,}032
\end{equation}
parameters to the base predictor.

The matrices $U_j$ and $V_j$ are initialized from
standard normal distributions. Gate weights and
biases are initialized to zero, so the conditional
model initially reproduces the base logits.
Additional tensors use an independent CPU random
number generator seeded with the training seed
plus 1,000,003, preserving the base-model
initialization and dropout random-number state.

\paragraph{Training and scoring.}
All variants sum seven equally weighted code
cross-entropies during training and the corresponding
log-probabilities during retrieval. Conditional
distributions are normalized over the full
vocabulary at each level, without a reachable-code
mask. Candidates are not excluded because their
codes lack positive training support.

\subsection{Training Configurations}

The internal comparisons use the same molecular
features, frozen tokenizer for each seed, and
positive training associations. Optimization uses
Adam with learning rate $10^{-4}$, zero weight decay,
$\beta=(0.9,0.999)$, $\epsilon=10^{-8}$, and batch
size up to 4,096.

Validation is performed every 94 updates.
Checkpoints are selected by validation R@10,
breaking ties by R@50, then MRR, and retaining the
earliest checkpoint for remaining ties.
Appendix~\ref{app:ablations} reports the controlled positive-hiding experiment.

These fixed internal settings are distinct from
the validation-selected headline configurations,
which are documented in
Appendix~\ref{app:transfer}.
Training used NVIDIA A100 40\,GB GPUs.

\subsection{Pfam Annotation and Code-Structure Analysis}
\label{app:code-visualization}

\paragraph{Annotation population.}
We use UniProt Pfam cross-references for the frozen
catalogue, retrieved on 10 September 2026.
Each accession--Pfam membership is counted once;
domain positions and repeated-domain copy counts
are not available in this analysis.

Of the 20,431 catalogue proteins, 19,457 (95.23\%)
have at least one Pfam annotation, 974 have none,
and 8,389 have multiple distinct annotations.
The analysis retains all 8,953 observed Pfam entries
and 34,956 memberships. Missing cross-references
are not interpreted as an absence of domains.
Pfam display names are obtained through the InterPro
API without changing the frozen memberships.
Pfam annotations are not used to train the tokenizer
or retrieval model.

\paragraph{Composition of displayed groups.}
For each displayed code group,
Figure~\ref{fig:code-semantics} identifies
its most frequent Pfam, with Pfam-ID order
breaking frequency ties. Proteins are divided
into three disjoint categories: those carrying
this Pfam, those carrying other Pfams but not
this Pfam, and those without annotations.

All group members are included. Assigning a
protein to a display category does not remove
any of its additional Pfam memberships from
the statistical analysis.

\paragraph{Within-Pfam code concentration.}
For tokenizer seed $s$, semantic level $j$,
Pfam $f$, and code $c$, let $n_{sjfc}$ count
proteins annotated with $f$ and assigned to $c$.
The concentration of $f$ at this level is
\begin{equation}
C_{sjf}
=
\frac{\max_c n_{sjfc}}{n_f},
\qquad
n_f=\sum_c n_{sjfc}.
\end{equation}

Figure~\ref{fig:code-semantics} reports
five-seed mean concentrations for the twelve
most prevalent Pfams, with Pfam-ID order
breaking frequency ties. These rows cover
3,328 unique proteins. Because proteins may
carry multiple Pfams, row populations overlap
and should not be summed.

Concentration measures how strongly a Pfam
is assigned to a single code, rather than a
one-to-one correspondence between codes and
families. Codebook sizes differ across levels,
so concentration values do not provide a
capacity-controlled ranking of semantic quality.

\paragraph{Protein-pair correspondence.}
Let $A_p$ be the set of Pfams assigned to
protein $p$, and let $\mathcal A$ contain
catalogue proteins with at least one annotation.
For seed $s$ and level $j$, we calculate
\begin{equation}
R_{sj}
=
\frac{
\sum_{\substack{p<q\\p,q\in\mathcal A}}
\mathbf{1}[c_{sj}(p)=c_{sj}(q)]
\mathbf{1}[A_p\cap A_q\neq\varnothing]
}{
\sum_{\substack{p<q\\p,q\in\mathcal A}}
\mathbf{1}[c_{sj}(p)=c_{sj}(q)]
}.
\end{equation}

Each unordered protein pair contributes once,
irrespective of the number of shared Pfams.
The catalogue reference is the corresponding
fraction across all annotated protein pairs,
without restricting them to a shared code:
0.4742\%. This is an observed background
fraction, not a simulated null distribution.

\begin{figure}[!htb]
\centering
\includegraphics[width=\linewidth]
{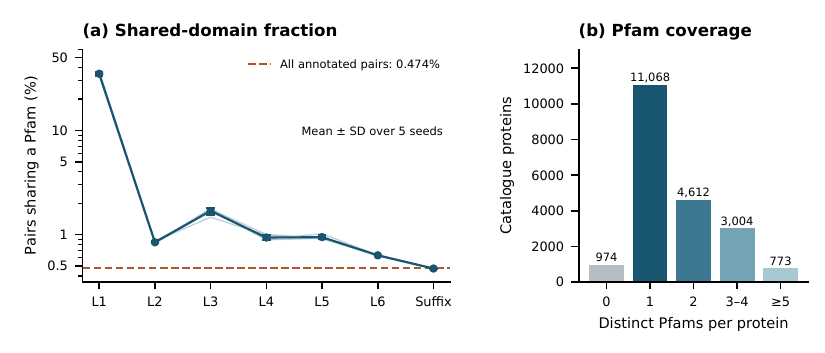}
\caption{
\textbf{Pfam correspondence across code levels
and catalogue annotation coverage.}
(a) Fraction of annotated protein pairs sharing
a code that also share at least one Pfam.
Thin lines show five tokenizer seeds; markers
and error bars indicate their mean and sample
standard deviation. The vertical axis is
logarithmic. The dashed line shows the fraction
among all annotated catalogue pairs.
The suffix resolves identifier collisions
and is not a semantic level.
(b) Catalogue protein counts by number of
distinct Pfam annotations; zero indicates
missing cross-references.
}
\label{fig:pfam-levels}
\end{figure}

Across tokenizer seeds 13, 37, 73, 101, and 211,
the first-level statistic is
$34.98\%\pm1.34\%$ (mean $\pm$ sample SD).
Figure~\ref{fig:pfam-levels} reports all semantic
levels and the collision suffix. Differences
in codebook sizes, group sizes, and eligible
pair populations should be considered when
comparing levels. These statistics assess
annotation correspondence; the retrieval
contribution of protein--code assignments is
evaluated separately in
Appendix~\ref{app:code-assignment}.

\section{Code Probabilities and Exact Catalogue Scoring}
\label{app:math}
Let $c(p)$ denote the complete seven-code identifier of protein $p$, and let $Q(c\mid x)$ denote the code distribution conditioned on molecule $x$. The tokenizer and identifier assignments remain fixed during molecular predictor training.
\subsection{Identifier and Probability Definitions}
The six semantic codes may be shared by multiple proteins. A fixed suffix distinguishes identities within each collision group, making complete identifiers unique within the encoded protein universe. Let $\mathcal P$ be the candidate catalogue and $\mathcal C=\prod_{j=1}^{7}\{0,\ldots,K_j-1\}$ the identifier space. GenFish defines
\begin{equation}
Q(c\mid x)=\prod_{j=1}^{7}P_j(c_j\mid x,c_{j-1}),\qquad c_0=\varnothing,
\label{eq:code-joint}
\end{equation}
where levels 1 and 7 omit predecessor conditioning. Each factor is normalized over its full vocabulary, so $Q$ sums to one over $\mathcal C$.
\subsection{Training Objective and Exact Scoring}
Summing seven equally weighted code cross-entropies gives the positive-pair loss $-\log Q(c(p)\mid x)$. The objective uses recorded positives without assigning experimental negative labels to unmeasured pairs. At retrieval time, each candidate is evaluated along its stored identifier; the sum of normalized code log-probabilities is the exact log-likelihood in Eq.~\ref{eq:code-joint}. Scoring requires neither a generated prefix nor knowledge of the query's true target.

\section{External Comparators and Training Configurations}
\label{app:transfer}

We compare GenFish with four neural methods and three
simple baselines using the same retained positive
associations and candidate catalogue. DrugCLIP, HypSeek,
and LigUnity are adapted to frozen ESMC-600M protein
features and Morgan-2048 molecular fingerprints.
ConPLex retains its native ProtBERT-based projection
architecture.

Prepared binding sites are unavailable for most
catalogue proteins, so pocket-based encoders are
replaced with sequence-feature towers. The adaptations
below retain selected components of the published
methods while specifying changes to inputs,
architecture, and objectives. The comparisons therefore
evaluate complete training configurations rather than
scoring functions alone.

\subsection{DrugCLIP}
\label{app:drugclip}

Our DrugCLIP adaptation uses dual encoders,
$L_2$-normalized embeddings, inner-product scoring,
and a bidirectional multi-positive contrastive
objective, following the retrieval formulation of
\citet{drugclip2023}. The protein tower operates
on frozen ESMC-600M features, while the molecular
tower uses Morgan-2048 fingerprints.
Architecture and optimization settings are given
in Appendix~\ref{app:headline-config}.

\subsection{HypSeek}
\label{app:hypseek}

The HypSeek adaptation uses a hyperbolic
representation based on \citet{hypseek2025}.
With curvature fixed to one, tower outputs
$v$ are lifted as
\begin{equation}
\operatorname{exp}_0(v)
=
\left(
\cosh\|v\|,
\frac{\sinh\|v\|}{\|v\|}v
\right),
\end{equation}
using the continuous limit at $v=0$.
Retrieval uses the temperature-scaled inner
product of the spatial components. Tangent
norms are capped so that spatial norms remain
within $[0,1]$.

The sequence towers and contrastive objective
follow the DrugCLIP adaptation.
The adapted objective omits the affinity-ordered
Plackett--Luce and entailment-cone terms.
The binary training view does not provide
within-assay affinity orderings for the
affinity-ordered component.

\subsection{LigUnity}
\label{app:ligunity}

The LigUnity adaptation retains cosine scoring,
bidirectional in-batch contrastive training,
and ligand-to-target and target-to-target
refinement branches \citep{ligunity2025}.
Ligand-to-target aggregation averages molecular
tower representations of each target's training
ligands. Two-hop target aggregation supplies
neighboring ligand profiles, including for
targets without direct training support.

Pocket encoders are replaced with sequence
towers. Each target connects to its 16 nearest
training-supported neighbors by frozen
protein-feature cosine similarity, replacing
pocket-alignment edges. Aggregation uses uniform
averaging instead of learned attention, and
the affinity-ordered pharmacophore-ranking
loss is omitted.

Refinement gates are initialized to zero.
Shared submodules are initialized before the
additional branches to preserve their initial
parameters across matched configurations.

\subsection{ConPLex}
\label{app:conplex}

We retain the official SimpleCoembedding
architecture of \citet{conplex2023}.
Morgan-2048 molecular inputs and frozen
ProtBERT-1024 protein features each pass through
a single linear projection followed by ReLU,
producing 1,024-dimensional representations
scored by cosine similarity.
Both projections are initialized afresh,
without a supervised interaction checkpoint.

ProtBERT features use mean pooling excluding
special tokens. Sequences are divided into
nonoverlapping windows of at most 1,022 residues,
with window representations aggregated by
residue count. Protein identities and
full-sequence hashes are checked against
the catalogue. Unlike the other neural
methods, ConPLex retains ProtBERT rather
than ESMC features.

Training uses binary cross-entropy on the
nonnegative cosine score with numerical
clamping. The external DUDE decoy-contrastive
stage is omitted. Each update contains
4,096 positive pairs and an equal number
of sampled unknown pairs.
The unknown-pair sample is fixed for each
seed, excludes each molecule's known training
positives, and draws targets only from the
training-supported set. Validation and test
labels are not used in sampling.
These unknown pairs provide negative training
labels but are not experimentally verified
negatives.

Optimization uses AdamW with weight decay
0.01 and cosine warm restarts every ten
positive-data passes. The 22,560-update
ceiling corresponds to 242.39 positive-data
passes, approximately 1\% more than the
240 passes of the ESMC-based methods.
Update counts are controlled, but pair
evaluations and computational cost are
not identical.

We search learning rates
$\{5\times10^{-5},10^{-4},2\times10^{-4}\}$
using seeds 13, 37, and 73.
Validation selects $10^{-4}$; seeds 101 and
211 use this setting without retuning.
All five selected checkpoints are reused
for measured-panel and similarity-stratified
evaluation.

Cached matrix scores agree with sampled
official pairwise forward scores within
$4.2\times10^{-6}$ absolute error.
The scorer retains the official zero-norm
convention for queries whose ReLU projections
are zero. All seeds are included, including
the variation in enzyme AP
($0.7035\pm0.1182$).

\subsection{Simple Baselines}
\label{app:simple-baselines}

All three baselines use the same deduplicated
training-positive associations and frozen
candidate catalogue.

\paragraph{Target frequency.}
Each protein is scored by its number of
distinct training ligands. This ranking is
independent of the query molecule.

\paragraph{Weighted k-NN.}
For each query, we retrieve the globally
nearest $k$ training molecules using Tanimoto
similarity on binary radius-two Morgan-2048
fingerprints. A target's score is the sum
of similarities from neighbors carrying
that target label, without target-frequency
correction. Targets receiving no votes
have score zero.

Validation selects $k=50$ from
$\{1,3,5,10,20,50,100,200\}$ by R@10,
then R@50 and MRR; remaining ties favor
smaller $k$. Neighbor ties use molecular
identifier order, and target-score ties
use the frozen catalogue order.

\paragraph{SEA.}
SEA (Morgan) reimplements the similarity
ensemble formulation
\citep{sea2007,keiser2009}.
For each target, the raw score $s$ sums
query-to-training-ligand Tanimoto similarities
strictly above 0.3. It is standardized
against a training-only background:
\begin{equation}
z=\frac{s-an}{bn^c},
\end{equation}
where $n$ is the target's training-ligand
count and the fitted constants are
$a=0.0006492743$,
$b=0.01237402$, and
$c=0.6397422$.
Targets are ranked by the negative logarithm
of the Gumbel-tail probability.
Targets without training ligands receive
score zero.

The similarity threshold follows the
authors' ECFP4 configuration, while
fingerprints and background fitting use
the present benchmark.
Background fitting uses 512 training
queries, 16 random subsets per query,
and 20 logarithmically spaced set sizes.
A separate set of 128 training queries
is used to diagnose calibration without
tuning. The tail probabilities describe
the approximate similarity background,
not biological binding probabilities.

The same fixed baseline scores are used
for full-catalogue retrieval,
measured-panel AP, and similarity-stratified
analysis.

\subsection{Headline Neural Configurations and Selection}
\label{app:headline-config}

All headline neural methods use Morgan-2048
molecular inputs and the same 381,225
positive training associations.
DrugCLIP, HypSeek, LigUnity, and GenFish
use frozen ESMC-600M protein features;
ConPLex uses the architecture and inputs
specified above.

For the ESMC-based methods, the shared
molecular trunk has two 512-wide ReLU
layers. The three adapted comparators
project molecular representations to
256 dimensions and use a protein tower
with one 512-wide hidden layer and a
256-dimensional output.
GenFish uses seven code heads and the
conditional modules in
Appendix~\ref{app:model-contract}.
LigUnity additionally uses the refinement
branches described above.

The four ESMC-based methods use batches
of up to 4,096 positive associations,
retain the final partial batch, and train
for at most 22,560 updates.
Validation is performed every 94 updates.
Weight decay is zero, with optimizer
parameters $\beta=(0.9,0.999)$ and
$\epsilon=10^{-8}$.

The contrastive objectives average
forward and reverse multi-positive
log-sum-exp losses, using known training
positives in the batch and up to 2,048
additional training-supported candidates.
GenFish sums seven equally weighted
code cross-entropies and does not use
a contrastive temperature.
Selected configurations are listed in
Table~\ref{tab:headline-config}.

\begin{table}[t]
\centering
\small
\setlength{\tabcolsep}{4pt}
\caption{
Validation-selected headline configurations.
Temperature applies only to contrastive
objectives. Parameter counts refer to
the trainable predictors and exclude
frozen protein feature extractors.
}
\label{tab:headline-config}
\begin{tabular}{@{}lrrrrl@{}}
\toprule
Method & Learning rate & Dropout &
Temperature & Parameters & Optimizer \\
\midrule
DrugCLIP (adapted)
& $5\times10^{-5}$ & 0.1 & 0.03
& 2,164,736 & AdamW \\
HypSeek (adapted)
& $10^{-4}$ & 0.1 & 0.03
& 2,164,736 & AdamW \\
LigUnity (adapted)
& $5\times10^{-5}$ & 0.1 & 0.07
& 4,525,570 & AdamW \\
ConPLex (no-DUDE)
& $10^{-4}$ & 0.0 & --
& 3,147,776 & AdamW \\
GenFish
& $10^{-4}$ & 0.0 & --
& 3,887,864 & Adam \\
\bottomrule
\end{tabular}
\end{table}

Configurations are selected using
three-seed mean validation performance
over seeds 13, 37, and 73.
Selection prioritizes R@10, then R@50,
then MRR, with configuration identifier
breaking remaining ties.
Seeds 101 and 211 are subsequently
trained without retuning.

Within each run, checkpoints follow
the same validation metric ordering,
with the earliest checkpoint retained
for exact ties. Measured-panel and
similarity-stratified evaluations reuse
these catalogue-selected checkpoints.
Selected update counts are provided
in the supplementary CSV; no headline
run selected the final update ceiling.

\subsection{Similarity-Stratified Retrieval}
\label{app:hard-band}

We stratify the 198 test molecules by
their maximum Morgan-fingerprint
Tanimoto similarity to a training ligand.
The groups contain 43 molecules below
0.4, 79 in $[0.4,0.6)$, and 76 at
or above 0.6.
Table~\ref{tab:similarity-results}
reports all eight methods.
Predictions and checkpoints are reused
without refitting or reselection.

\begin{table}[!tp]
\centering
\footnotesize
\setlength{\tabcolsep}{3.5pt}
\caption{Full-catalogue retrieval stratified by maximum
training-ligand Tanimoto similarity.
All methods reuse the scores in Table~\ref{tab:retrieval-results},
without refitting. Neural entries report five-seed means
$\pm$ sample SD; other entries are single results.
Higher is better for all metrics ($\uparrow$);
bold marks the highest mean within each metric and similarity group.}
\label{tab:similarity-results}

\begin{tabular}{@{}l r r r@{}}
\toprule
\multicolumn{4}{@{}l}{\textbf{Retrieval by training-ligand similarity}}\\
\multicolumn{4}{@{}l}{\emph{Maximum Morgan-fingerprint Tanimoto similarity; median 0.511}}\\
Method & $[0,0.4)$, $n=43$ & $[0.4,0.6)$, $n=79$ & $[0.6,1]$, $n=76$\\
\midrule

\multicolumn{4}{@{}l}{\textit{R@10} $\uparrow$}\\
Target frequency
& $0.0000$ & $0.0248$ & $0.0135$\\
Weighted k-NN
& $0.0777$ & $\mathbf{0.4332}$ & $0.4130$\\
SEA (Morgan)
& $0.0667$ & $0.4059$ & $0.3411$\\
DrugCLIP (adapted)
& $0.0527\pm0.0066$ & $0.3724\pm0.0149$ & $0.3539\pm0.0199$\\
HypSeek (adapted)
& $0.0342\pm0.0141$ & $0.3312\pm0.0197$ & $0.3133\pm0.0339$\\
LigUnity (adapted)
& $0.0847\pm0.0153$ & $0.3910\pm0.0170$ & $0.4017\pm0.0136$\\
ConPLex (no-DUDE)
& $0.0514\pm0.0151$ & $0.3562\pm0.0158$ & $0.3016\pm0.0363$\\
GenFish
& $\mathbf{0.0905\pm0.0152}$
& $0.4143\pm0.0095$
& $\mathbf{0.4371\pm0.0141}$\\
\midrule

\multicolumn{4}{@{}l}{\textit{R@50} $\uparrow$}\\
Target frequency
& $0.1197$ & $0.1308$ & $0.1516$\\
Weighted k-NN
& $0.0874$ & $0.4866$ & $0.5354$\\
SEA (Morgan)
& $0.0667$ & $0.4777$ & $0.5318$\\
DrugCLIP (adapted)
& $0.0815\pm0.0177$ & $0.4394\pm0.0403$ & $0.4662\pm0.0551$\\
HypSeek (adapted)
& $0.0546\pm0.0150$ & $0.3961\pm0.0300$ & $0.4437\pm0.0759$\\
LigUnity (adapted)
& $0.1400\pm0.0414$ & $0.4560\pm0.0209$ & $0.5217\pm0.0228$\\
ConPLex (no-DUDE)
& $0.1185\pm0.0121$ & $0.5237\pm0.0133$ & $0.5999\pm0.0444$\\
GenFish
& $\mathbf{0.1496\pm0.0249}$
& $\mathbf{0.5503\pm0.0098}$
& $\mathbf{0.6971\pm0.0274}$\\
\midrule

\multicolumn{4}{@{}l}{\textit{MRR} $\uparrow$}\\
Target frequency
& $0.0091$ & $0.0303$ & $0.0551$\\
Weighted k-NN
& $0.0972$ & $0.4584$ & $0.5698$\\
SEA (Morgan)
& $0.0625$ & $0.4188$ & $0.4499$\\
DrugCLIP (adapted)
& $0.0691\pm0.0187$ & $0.4192\pm0.0090$ & $0.6048\pm0.0293$\\
HypSeek (adapted)
& $0.0432\pm0.0179$ & $0.3617\pm0.0317$ & $0.5130\pm0.0396$\\
LigUnity (adapted)
& $0.0898\pm0.0144$ & $0.4346\pm0.0116$ & $0.6211\pm0.0118$\\
ConPLex (no-DUDE)
& $0.0560\pm0.0142$ & $0.3565\pm0.0702$ & $0.4176\pm0.0287$\\
GenFish
& $\mathbf{0.1150\pm0.0179}$
& $\mathbf{0.4846\pm0.0187}$
& $\mathbf{0.6377\pm0.0350}$\\
\bottomrule
\end{tabular}
\end{table}

For the below-0.4 group, we compare
GenFish with LigUnity, the strongest
overall neural comparator on R@10
and MRR. The mean differences are
$+0.0059$ for R@10,
$+0.0096$ for R@50, and
$+0.0252$ for MRR.
The corresponding paired intervals are
$[-0.0182,+0.0306]$,
$[-0.0650,+0.0812]$, and
$[-0.0039,+0.0578]$.

GenFish leads in two of five seed pairs
for R@10, three for R@50, and all five
for MRR. All three intervals include
zero, so the higher means do not
establish a statistically resolved
ordering in this small group.

\subsection{Statistical Comparisons}

All reported comparison intervals use 10,000
paired molecule-bootstrap resamples. Seed-specific
metric values are first averaged for each molecule;
the same sampled molecule indices are then applied
to both methods.

The low-similarity comparisons with LigUnity use
pointwise 95\% percentile intervals, defined by
the 2.5th and 97.5th bootstrap percentiles.
Seed-wise win counts are reported separately.

The two prespecified R@50 comparisons against
target frequency and weighted k-NN use
Bonferroni-adjusted intervals targeting 95\%
family-wise coverage. Each interval uses the
1.25th and 98.75th bootstrap percentiles.
The separately conducted SEA comparison uses
a pointwise 95\% interval without multiplicity
adjustment.

These intervals quantify query-sampling uncertainty
conditional on the fitted models and baseline
configurations; they do not incorporate
training-seed uncertainty.
\section{Controlled Experiments}
\label{app:ablations}
This appendix gives the full protocol for the false-negative intervention and
the protein--code assignment permutation; the former is summarized in the
main text, while the latter is reported here as a supplementary control.
\subsection{Controlled Negative-Supervision Intervention}
\label{app:negative-supervision}

\paragraph{Hidden-positive construction.}
Using split seed 20260920, we reserve at
least one positive association per molecule
and originally supported protein from the
381,225 training positives.
Among the remaining 82,581 eligible pairs
whose targets belong to the evaluation
catalogue, we hide 41,290 pairs
(half, rounded down).
This leaves 339,935 visible positives
while preserving molecule and target
training support.

The hidden set covers 31,895 training-seen
molecules and 1,138 targets.
It represents 10.83\% of all training
associations; the 50\% hiding fraction
applies only to the eligible subset.
Validation and test labels remain unchanged.

\paragraph{Contrastive interventions.}
Standard training uses the visible positives
with the adapted DrugCLIP model's
bidirectional multi-positive contrastive loss.

Oracle masking removes hidden-positive
cells from the forward and active reverse
competition denominators without restoring
their positive labels.
Visible-positive numerators, reverse-anchor
eligibility, candidate sets, and loss-averaging
denominators remain unchanged.

Matched random masking removes the same
number of cells per molecule and
reverse-active status, excluding all
originally known positives, including
the hidden ones.
It preferentially matches target-support
bins of 1--4, 5--19, 20--99, 100--499,
and at least 500 training associations.
When a bin lacks eligible cells, selection
falls back to the nearest available
log-support degree with the same
reverse-active status.

Fixed pair priorities make the
batch-conditional selection deterministic.
Masking does not introduce replacement
candidates. Oracle masking uses withheld
ground truth to remove false-negative
competition, while random masking controls
for a matched reduction in competition.

\paragraph{Matched training.}
The three contrastive treatments and
GenFish use seeds 13, 37, and 73,
4,096-positive batches, matched
positive-example schedules, and
22,560 updates.
The contrastive treatments additionally
share initialization, candidate schedules,
and dropout streams.

Previously selected model-specific
settings remain fixed.
DrugCLIP uses AdamW with learning rate
$5\times10^{-5}$, dropout 0.1,
and temperature 0.03.
GenFish uses Adam with learning rate
$10^{-4}$, dropout 0, and its frozen
tokenizer. Both use zero weight decay.
Architecture and headline configurations
are described in
Appendices~\ref{app:model-contract}
and~\ref{app:transfer}.

All treatments are evaluated using their
final checkpoints, without checkpoint
selection on hidden-positive or test
performance. The GenFish extension was
specified after the contrastive results
and before its training runs.

\paragraph{Evaluation endpoints.}
Both endpoints use molecule-macro R@50
over the same 20,431-protein catalogue.

Hidden-positive recovery evaluates the
withheld targets of the 31,895 affected
training-seen molecules. Visible positives
remain in the ranked candidate catalogue
but are not counted as hidden positives.
Held-out retrieval evaluates the original
198 test molecules using their unchanged
positive labels.

Figure~\ref{fig:negative-supervision}
reports means and sample standard
deviations over three seeds.

\paragraph{Uncertainty and supplementary effects.}
Effect intervals use seed-averaged
per-molecule metric values and 10,000
paired molecule-bootstrap resamples.
Bonferroni-adjusted intervals target
95\% family-wise coverage separately
for two families of four contrasts:
oracle versus standard and random
masking at both endpoints; and GenFish
versus standard and oracle-masked
training at both endpoints.
The intervals condition on the fitted
models and the fixed hidden-positive
split.

Section~\ref{sec:negative-supervision}
reports the held-out oracle effects.
For hidden-positive recovery, oracle
masking improves over standard and
random-masked training by 1.91 and
2.09 percentage points, with adjusted
intervals $[1.78,2.04]$ and
$[1.95,2.23]$, respectively.
All four oracle contrasts are positive
in every seed.

For held-out retrieval, GenFish exceeds
standard and oracle-masked contrastive
training by 19.29 and 17.19 percentage
points, with adjusted intervals
$[15.13,23.79]$ and $[13.00,21.65]$,
respectively.

\subsection{Protein--Code Assignment Permutation}
\label{app:code-assignment}

The frequency-matched null permutes complete seven-code tuples only between proteins with equal training-association degree. Thus the relation-weighted frequency of every code value is unchanged at every level, while the assignment of tuples to proteins is altered. Across the three tokenizer seeds, $97.4$--$97.7\%$ of tuples move. We retrain Independent, Factorized, and GenFish under the original and permuted assignments using the same 120-epoch recipe, seeds 13, 37, and 73, and fixed 20,431-protein catalogue. Molecule--protein labels, model architecture, optimization schedule, and evaluation protocol are otherwise unchanged.

Under this controlled recipe, the original assignment exceeds the frequency-matched null by 2.62 percentage points in R@10 and 5.56 points in R@50, with the original assignment preferred for all three seeds. The control therefore tests whether the protein--code correspondence carries information beyond code frequency. It is a retrieval control rather than a claim that the quantizer alone causes all observed biological structure; the independent Pfam analysis in Section~\ref{sec:code-semantics} provides complementary evidence.

\end{document}